\documentclass{article} 
\usepackage[final,nonatbib]{neurips_data_2022}
\usepackage[numbers]{natbib}

\usepackage[utf8]{inputenc} 
\usepackage[T1]{fontenc}    
\usepackage{hyperref}       
\usepackage{url}            
\usepackage{booktabs}       
\usepackage{amsfonts}       
\usepackage{nicefrac}       
\usepackage{microtype}      
\usepackage{xcolor}         

\usepackage{graphicx}
\usepackage{subfigure}

\usepackage{amsmath,amsthm}
\usepackage[ruled,vlined]{algorithm2e}
\usepackage[utf8]{inputenc}
\usepackage[hang,flushmargin]{footmisc}
\usepackage{threeparttable,multirow}
\usepackage{tikz}
\usetikzlibrary{tikzmark,calc}
\usepackage{verbatim}
\usepackage{wrapfig,lipsum}
\usepackage{tablefootnote}

\newcommand{\A}{\mathcal A}

\newcommand{\D}{\mathcal D}
\newcommand{\E}{\mathbb E}

\newcommand{\R}{\mathbb R}
\newcommand{\W}{\mathcal W}
\newcommand{\X}{\mathcal X}

\usepackage{scalerel,mathtools,color}
\let\svsqrt\sqrt
\newsavebox\Nsqrt
\def\sr#1{\ThisStyle{%
		\savebox\Nsqrt{\scalebox{.5}[1]{$\SavedStyle\svsqrt{\phantom{\cramped{#1#1}}}$}}%
		\ooalign{\usebox{\Nsqrt}\cr\kern.2pt\usebox{\Nsqrt}\cr\hfil$\SavedStyle\cramped{#1}$}}}

\usepackage{enumitem}
\def\*#1{\mathbf{#1}}

\newcommand{\res}[2]{#1$\pm$#2}

\title{NAS-Bench-360: Benchmarking \\Neural Architecture Search on Diverse Tasks}

\author{%
  Renbo Tu\thanks{Equal contribution} \\
  University of Toronto \\
  \texttt{renbo.tu@mail.utoronto.ca} \\
    \And
   Nicholas Roberts$^\ast$ \\
  University of Wisconsin \\
   \texttt{nick11roberts@cs.wisc.edu} \\
   \AND
   Mikhail Khodak  \\
  Carnegie Mellon University\\
  \texttt{khodak@cmu.edu} \\
    \And
     Junhong Shen  \\
  Carnegie Mellon University\\
  \texttt{junhongs@andrew.cmu.edu} \\
    \AND
  Frederic Sala\\
 University of Wisconsin \\
  \texttt{fsala@wisc.edu} \\
   \And
  Ameet Talwalkar\\
   Carnegie Mellon University \\
  \texttt{talwalkar@cmu.edu} \\
}

\begin{document}

\maketitle
\setcounter{footnote}{0}

\begin{abstract}
Most existing neural architecture search (NAS) benchmarks and algorithms prioritize well-studied tasks, e.g. image classification on CIFAR or ImageNet.
This makes the performance of NAS approaches in more diverse areas poorly understood.
In this paper, we present {\bf NAS-Bench-360}, a benchmark suite to evaluate methods on domains beyond those traditionally studied in architecture search, and use it to address the following question: {\em do state-of-the-art NAS methods perform well on diverse tasks?}
To construct the benchmark, we curate ten tasks spanning a diverse array of application domains, dataset sizes, problem dimensionalities, and learning objectives. 
Each new task is carefully chosen to interoperate with modern convolutional neural network (CNN) search methods while being far-afield from their original development domain. 
To speed up and reduce the cost of NAS research, for two of the tasks we release the precomputed performance of 15,625 architectures comprising a standard CNN search space.
Experimentally, we show the need for more robust NAS evaluation of the kind NAS-Bench-360 enables by showing that several modern NAS procedures perform inconsistently across the ten tasks, with many catastrophically poor results.
We also demonstrate how our benchmark and its associated precomputed results will enable future scientific discoveries by testing whether several recent hypotheses promoted in the NAS literature hold on diverse tasks. NAS-Bench-360 is hosted at \url{https://nb360.ml.cmu.edu/}.
\end{abstract}


\section{Introduction}\label{sec:intro}

Neural architecture search (NAS) aims to automate the design of deep neural networks, ensuring performance on par with hand-crafted architectures while reducing human labor devoted to tedious architecture tuning \citep{elsken2019nas}. 
With the growing number of application areas of ML, and thus of use-cases for automating it, NAS has experienced an intense amount of study in well-established machine learning domains, with significant progress in search space design~\citep{zoph2018nas,liu2019darts,cai2019proxyless}, search efficiency~\citep{pham2018enas}, and search algorithms~\citep{xu2020pcdarts,li2021gaea,white2021bananas}.
Notably, the field has largely been dominated by methods designed for and evaluated on benchmarks in computer vision \citep{liu2019darts,ying2019nasbench101,dong2020nasbench201}, yet the use of NAS techniques may be especially impactful in under-explored or under-resourced domains where less is known about useful architecture design patterns. 
There have been a few recent efforts to diversify these benchmarks to settings such as vision-based transfer learning \citep{duan2021transnas} and speech and language processing \cite{mehrotra2021asr,klyuchnikov2020nlp};
however, evaluating NAS methods on such well-studied tasks using traditional CNN search spaces does not give a good indication of their utility on more far-afield applications, which have often necessitated the design of custom neural operations \citep{cohen2018spherical,li2021fno}.

We make progress towards studying NAS on more diverse tasks by introducing a suite of benchmark datasets drawn from various data domains that we collectively call {\bf NAS-Bench-360}.
This benchmark consists of an organized setup of ten suitable datasets that represent diverse application domains, dataset sizes, problem dimensionalities, and learning objectives.
We also include a standard image classification task as a baseline point of comparison, as many new methods continue to be designed for that setting. 
Note that the core component of NAS-Bench-360 is {\em not} a typical NAS benchmark, which often involves precomputing all architectures in some fixed search space.
In contrast, our contribution is explicitly intended to be agnostic of the search space being used, as different search spaces may work well for different tasks. 
Thus NAS-Bench-360 is a task-oriented NAS benchmark with the intended use-case of evaluating NAS method and search space pairs on a wide variety of domains.
However, to aid research, three of our tasks---for two of which we contribute the precompute---do come with trained architectures from the NAS-Bench-201 search space \citep{dong2020nasbench201}.

\begin{figure}[!t]
	
	\centering
	\includegraphics[width=0.49\linewidth]{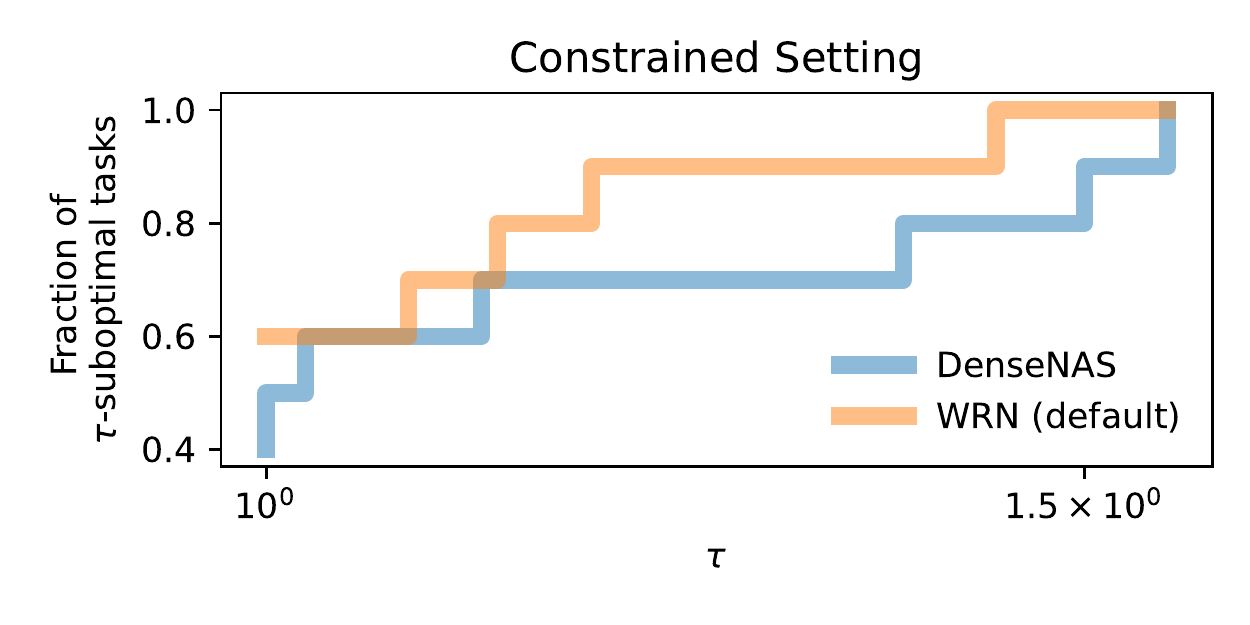}
	\hfill
	\includegraphics[width=0.49\linewidth]{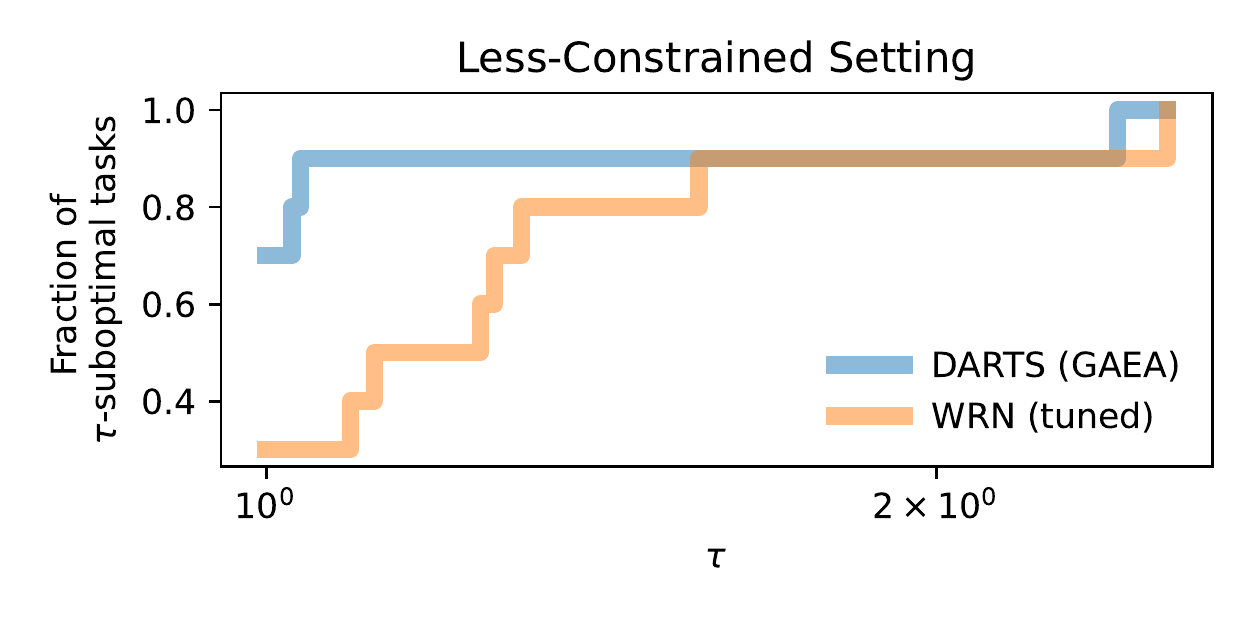}
	\vspace{-3mm}
	\caption{
		Performance profiles on NAS-Bench-360 comparing NAS methods (blue) to a fixed CNN (orange), specifically a Wide ResNet (WRN) \citep{zagoruyko2016wideresnet}. 
		Resource-constrained practitioners might be better off not using NAS (left), while less constrained practitioners can still benefit (right). 
		The y-axis is the fraction of tasks on which error is within a factor $\tau$ of the optimal method, i.e. higher is better. 
	}
	\label{fig:comparison}
	
\end{figure}

Experimentally, we demonstrate the usefulness of NAS-Bench-360 by performing a set of analyses evaluating whether the success of NAS in computer vision is indicative of strong performance on the much broader set of problems to which NAS can be applied.
Specifically, we report performance comparisons between NAS methods, investigate the validity of existing NAS hypotheses made solely on computer vision tasks, and extend an existing analysis of zero-cost proxies already-enabled by our benchmark~\cite{colin2022adeeperlook}. From these analyses, we arrive at the following conclusions:
\begin{itemize}[leftmargin=*,topsep=-1pt,noitemsep]\setlength\itemsep{2pt}
\item Resource-constrained practitioners may be better of using a fixed CNN rather than NAS (Figure~\ref{fig:comparison}).
\item NAS-Bench-201 analyses on computer vision tasks do not generalize to diverse tasks. 
\item Zero-cost proxies perform inconsistently on diverse tasks, corroborating earlier findings \cite{colin2022adeeperlook}. 
\end{itemize}

We have released all datasets, experiment code, precomputed models, seeds, and environments used in our experiments.\footnote{\url{https://github.com/rtu715/NAS-Bench-360}} 
Releasing our code, random seeds, and environments in the form of Docker containers assures reproducibility of all experimental results presented in this work and encourages the same level of reproducibility for future research performed using NAS-Bench-360.

\section{Related Work}\label{sec:related}
Benchmarks have been critical to the development of NAS in recent years.
This includes standard evaluation datasets and protocols, of which the most popular are the CIFAR-10 and ImageNet routines used by DARTS~\citep{liu2019darts}.
Another important type of benchmark has been tabular benchmarks such as NAS-Bench-101~\citep{ying2019nasbench101}, NAS-Bench-201~\citep{dong2020nasbench201}, NAS-Bench-1Shot1~\citep{zela2020nasbench1shot1}, and TransNAS-Bench-101~\citep{Duan2021TransNASBench101IT};
these benchmarks exhaustively evaluate all architectures in their search spaces, which is made computationally feasible by defining simple searched cells.
Consequently, they are less expressive than the DARTS cell \citep{liu2019darts}, often regarded as the most powerful search space in the cell-based regime.
Notably, the full NAS-Bench-360 benchmark is {\em not} intended to be a tabular benchmark, i.e. we do {\em not} evaluate every architecture from a fixed search space on all ten of our tasks; instead, the focus is on the organization of a suite of tasks for assessing both NAS algorithms and search spaces, which would necessarily be restricted by fixing a search space for a tabular benchmark. 
Pre-computing on an expansive search space such as DARTS, with $10^{18}$ possible architectures, is computationally intractable. 
Architectures found on lesser search spaces are most likely suboptimal: a vanilla Wide ResNet (WRN) outperforms all networks in the NAS-Bench-201 search space on CIFAR-100.
Nonetheless, we find that including precompute results for all of NAS-Bench-201 on two of our tasks is useful in evaluating various claims in the NAS literature centered on computer vision tasks.\looseness-1

While NAS methods and benchmarks have generally been focused on computer vision, recent work such as AutoML-Zero \citep{real2020automlzero} and XD-operations \citep{roberts2021xd} has started moving towards a more generically applicable set of tools for AutoML.
However, even more recent benchmarks that do go beyond the most popular vision datasets have continued to focus on well-studied tasks, including vision-based transfer learning~\citep{duan2021transnas}, speech recognition~\citep{mehrotra2021asr}, and natural language processing~\citep{klyuchnikov2020nlp}.
We aim to go beyond such areas to evaluate the potential of NAS to automate the application of ML in truly under-explored domains.
One analogous work to ours in the field of meta-learning is the Meta-Dataset benchmark of few-shot tasks \citep{triantafillou2020metadataset}, which similarly aimed to establish a wide-ranging set of evaluations for that field. For our inclusion of diverse tasks, we title our benchmark NAS-Bench-360 to resemble the idea of a 360-degree camera that covers all possible directions. 

\section{NAS-Bench-360: A Suite of Diverse and Practical Tasks}\label{sec:tasks}

In this section, we introduce the NAS setting targeted by our benchmark, our motivation for organizing a new set of diverse tasks as a NAS evaluation suite, and our task-selection methodology.
We report evaluations of specific algorithms on this new benchmark in the next section.

\subsection{Neural Architecture Search: Problem Formulation and Baselines}

For completeness and clarity, we first formally discuss the architecture search problem itself, starting with the extended hypothesis class formulation \cite{li2021gaea}.
Here the goal is to use a dataset of points $x\in\X$ to find parameters $\*w\in\W$ and $a\in\A$ of a parameterized function $f_{\*w,a}:\X\mapsto\R_{\ge0}$ that minimize the expectation $\E_{x\sim\D}f_{\*w,a}(x)$ for some test distribution $\D$ over $\X$;
here $\X$ is the input space, $\W$ is the space of model weights, and $\A$ is the set of architectures.
For generality, we do not require the training points to be drawn from $\D$ to allow for domain adaptation, as is the case for one of our tasks, and we do not require the loss to be supervised.
Note also that the goal here does not depend on computational or memory efficiency, which we do not focus on in our evaluations; our restriction is only that the entire pipeline can be run on an NVIDIA V100 GPU.

Notably, this formulation makes no distinction between the model weights $\*w$ and architectures $a$, treating both as parameters of a larger model.
Indeed, the goal of NAS may be seen as similar to model design, except now we include the design of an (often discrete) {\em architecture space} $\A$ such that it is easy to find an architecture $a\in\A$ and model weights $\*w\in\W$ whose test loss $\E_\D f_{\*w,a}$ is low using a search algorithm.
This can be done in a one-shot manner---simultaneously optimizing $a$ and $\*w$---or using the standard approach of first finding an architecture $a$ and then keeping it fixed while training model weights $\*w$ using a pre-specified algorithm such as stochastic gradient descent (SGD). This formulation divides NAS algorithms into two camps: one-shot, weight-sharing methods and non-weight-sharing ones such as random search, which operate by repeatedly sampling architectures and evaluating them.
The formulation also includes non-NAS methods by allowing the architecture search space to be a singleton.
When the sole architecture is a standard and common network such as WRN~\citep{zagoruyko2016wideresnet}, this yields a natural baseline with an algorithm searching for training hyperparameters, not architectures. 
For our empirical investigation, we compare the performance of state-of-the-art NAS approaches against that of the three baselines: WRN, PerceiverIO~\cite{jaegle2022perceiver}, and XGBoost~\cite{xgboost}.

\begin{table}[h!]
	\footnotesize
	\caption{Task metadata for NAS-Bench-360. Metrics are standardized such that lower is better.}
	\label{table-1}
	\centering
	\begin{tabular}{l@{\hspace{10pt}}lllll@{\hspace{4pt}}c}
		\toprule
		
		Task name   & Size  & Dim. & Type &  Learning objective & Metric & New to NAS?\\
		\midrule
		CIFAR-100 & \multirow{2}{*}{60K}  & \multirow{2}{*}{2D} & \multirow{2}{*}{Point} &  Classify natural images & \multirow{2}{*}{0-1 error} & no, widely \\
		\cite{krizhevsky2009cifar}&&&& into 100 classes && used \\

		\midrule
		Spherical & \multirow{2}{*}{60K}  &  \multirow{2}{*}{2D} & \multirow{2}{*}{Point} &  Classify spherically projected   & \multirow{2}{*}{0-1 error} &
		\multirow{2}{*}{\checkmark} \\
\cite{cohen2018spherical}&&&& images into 100 classes & \\
		
		\midrule
		NinaPro & \multirow{2}{*}{3956}  &  \multirow{2}{*}{2D}& \multirow{2}{*}{Point} &  Classify sEMG signals into  & \multirow{2}{*}{0-1 error} &
\multirow{2}{*}{\checkmark} \\
\cite{atzori2012building}&&&& 18 classes of hand gestures & \\
		\midrule
		\multirow{2}{*}{FSD50K} & \multirow{3}{*}{51K} &  \multirow{3}{*}{2D} & 
		Point & Classify sound events & \multirow{3}{*}{$1\hspace{-0.5mm}-\hspace{-0.5mm}\text{mAP}$} &
\multirow{3}{*}{\checkmark} \\  
\multirow{2}{*}{\cite{fonseca2017freesound}}&&&(multi-& in log-mel spectrograms & \\
&&&~label)& with 200 labels & \\
		\midrule
		Darcy Flow & \multirow{2}{*}{1100}  & \multirow{2}{*} {2D} & \multirow{2}{*}{Dense} &  Predict the final state of a fluid & \multirow{2}{*}{relative $\ell_2$} & no, used \\
\cite{li2021fno}&&&& from its initial conditions && in~\cite{roberts2021xd} \\
		\midrule
		\multirow{2}{*}{PSICOV} & \multirow{3}{*}{3606} &  \multirow{3}{*}{2D}  & \multirow{3}{*}{Dense} &  Predict pairwise distances & \multirow{3}{*}{$\text{MAE}_8$} & \multirow{2}{*}{no, used} \\
\multirow{2}{*}{\cite{adhikari2020fully}}&&&& between residuals from & & \multirow{2}{*}{in~\cite{roberts2021xd}}\\
&&&& pairwise sequence features & \\

		\midrule
\multirow{2}{*}{Cosmic} & \multirow{3}{*}{5250} &  \multirow{3}{*}{2D}  & \multirow{3}{*}{Dense} &  Predict probabilistic maps & \multirow{3}{*}{1 - AUROC} &
\multirow{3}{*}{\checkmark} \\
\multirow{2}{*}{\cite{zhang2020deepcr}}&&&& to identify cosmic rays  & \\
&&&& in telescope images  & \\

		\midrule 
ECG & \multirow{2}{*}{330K} &  \multirow{2}{*}{1D}  & \multirow{2}{*}{Point} &  Detecting atrial cardiac disease & \multirow{2}{*}{$1\hspace{-0.5mm}-\hspace{-0.5mm}\text{F1}$} &
\multirow{2}{*}{\checkmark} \\
\cite{clifford2017af}&&&& from ECG recordings & \\

		\midrule 
Satellite & \multirow{2}{*}{1M} &  \multirow{2}{*}{1D}  & \multirow{2}{*}{Point} &  Classify satellite image pixel time  & \multirow{2}{*}{0-1 error} &
\multirow{2}{*}{\checkmark} \\
\cite{petitjean2012satellite}&&&& series into 24 land cover types  & \\

		\midrule
\multirow{2}{*}{DeepSEA} & \multirow{3}{*}{250K} &  \multirow{3}{*}{1D} & 
Point & Predicting chromatin  & \multirow{3}{*}{$1\hspace{-0.5mm}-\hspace{-0.5mm}\text{AUROC}$} & no, used \\ 
\multirow{2}{*}{\cite{encode2004encode}}&&&(multi-&  and binding states  & & in \\
&&&~label)& of DNA sequences && \cite{zhang2021ambient,zhang2021automated} \\
		\bottomrule
	\end{tabular}
\end{table}

\subsection{Task Selection: Motivation and Methodology}
Curating a diverse, practical set of tasks for the study of NAS is our primary motivation behind this work. We observe that past NAS benchmarks focused on creating larger search spaces and more sophisticated search methods for neural networks. However, the utility of these search spaces and methods are only evaluated on canonical computer vision datasets. On a broader range of problems, whether these new methods can improve upon simple baselines remains an open question. This calls for the introduction of new datasets lest NAS research overfits to the biases of CIFAR-10 and ImageNet. By identifying these possible biases, future directions in NAS research can be better primed to suit the needs of practitioners and to increase the deployment of NAS.

Summarized in Table \ref{table-1}, NAS-Bench-360 consists of problems that are conducive to processing by convolutional neural networks, which includes a trove of applications associated with spatial and temporal data, spanning single and multiple dimensions. Most current NAS methods are not implemented to search for other types of architectures to process tabular data and graph data. Therefore, we have set this scope for our investigation.
During the selection of tasks, diversity is our primary consideration. We define the following axes of diversity to govern our task-filtering process: the first is problem dimensionality, including both 2D with matrix inputs and 1D with sequence inputs; the second is dataset size, for which our selection spans the scale from 1,000 to 1,000,000; the third is problem type , divisible into tasks requiring a singular prediction (point prediction) and multiple predictions (dense prediction); fourth and finally, diversity is achieved through selecting tasks from various learning objectives from applications of deep learning, where introducing NAS could improve upon the performance of handcrafted neural networks.

In lieu of providing raw data, we perform data pre-processing locally and store the processed data on a public Amazon Web Services S3 data bucket with download links available on our website. Our data treatment largely follows the procedure defined by the researchers who provided them. This enhances reproducibility by ensuring the uniformity of input data for different pipelines. Additional information about the datasets, pre-processing, and augmentation steps are described in the Appendix.


\section{Experimental design}

\begin{table}
	\centering
	\begin{threeparttable}
		\caption{
			Performance of NAS and baselines across NAS-Bench-360. Methods are divided into efficient methods (e.g. DenseNAS and fixed WRN) that take 1-10 GPU-hours, more expensive methods (e.g. DARTS and tuned WRN) that take 10-100+ GPU-hours, and specialized methods (Auto-DL and AMBER).
			All results are averages of three random seeds, and lower is better for all metrics. The best performing method is shown in \textbf{bold} and the best non-expert-designed method is \underline{underlined}.\looseness-1
		}
		\label{table-2}
		
		\footnotesize
		\begin{tabular}{llccccc}
			\toprule
			Search & Search & \multirow{2}{*}{CIFAR-100} & \multirow{2}{*}{Spherical}  & \multirow{2}{*}{Darcy Flow} & \multirow{2}{*}{PSICOV} & \multirow{2}{*}{Cosmic} \\
			space & algorithm &  &  &  &  & \\
			
			\midrule
			WRN & default &\underline{\res{23.35}{0.05}}& \res{85.77}{0.71}& \res{0.073}{0.001}& \res{3.84}{0.05}& \res{0.245}{0.02} \\
			DenseNAS & random & \res{25.49}{0.41}& \res{71.23}{1.65}&\res{0.071}{0.006}& \res{3.70}{0.06}& \res{0.309}{0.04}  \\
			DenseNAS & original & \res{25.98}{0.38} &\res{72.99}{0.95} & \res{0.100}{0.010}& \res{3.84}{0.15}& \res{0.383}{0.04} \\
			Perceiver IO & default & \res{70.04}{0.44} & \res{82.57}{0.19} & \res{0.240}{0.010} & \res{8.06}{0.06} & \res{0.485}{0.01} \\
			XGBoost & default & \res{84.83}{4.15} & \res{96.92}{0.02} & \res{0.085}{0.000} & n/a$^*$ & \res{0.232}{0.00} \\
			
			\midrule
			WRN & ASHA & \res{23.39}{0.01}  & \res{75.46}{0.40} & \res{0.066}{0.000}  &\res{3.84}{0.05} & \res{0.251}{0.02}  \\
			DARTS & GAEA &\res{24.02}{1.92}  & \underline{\bf\res{48.23}{2.87}}  & \underline{\res{0.026}{0.001}}  & \underline{\bf\res{2.94}{0.13}} & \underline{\res{0.229}{0.04}}  \\ 
			
			\midrule 		
			Auto-DL & DARTS &n/a& n/a&\res{0.049}{0.005}&\res{6.73}{0.73}& \res{0.495}{0.00}\\ 
			\midrule
			Expert & default & {\bf\res{19.39}{0.20}} & \res{67.41}{0.76} &\textbf{ \res{0.008}{0.001}} & \res{3.35}{0.14} & {\bf\res{0.127}{0.01}}  \\
			\toprule
			\toprule
			
			Search & Search & \multirow{2}{*}{NinaPro} & \multirow{2}{*}{FSD50K}  & \multirow{2}{*}{ECG} & \multirow{2}{*}{Satellite} & \multirow{2}{*}{DeepSEA} \\	
			space & algorithm & & & & & \\
			
			\midrule
			WRN & default & \underline{\bf\res{\,\,6.78}{0.26}} & \res{0.92}{0.001}&\res{0.43}{0.01}&\res{15.49}{0.03}& \res{0.40}{0.001} \\
			DenseNAS & random & \res{\,\,8.45}{0.56}&\underline{\bf\res{0.60}{0.001}}&\res{0.42}{0.01}&\res{13.91}{0.13}&\res{0.40}{0.001} \\
			DenseNAS & original & \res{10.17}{1.31}&\res{0.64}{0.002}&\res{0.40}{0.01}&\res{13.81}{0.69}&\res{0.40}{0.001} \\
			Perceiver IO & default & \res{22.22}{1.80} & \res{0.72}{0.002} & \res{0.66}{0.01} & \res{15.93}{0.08} & \res{0.38}{0.004} \\
			XGBoost & default & \res{21.90}{0.70} & \res{0.98}{0.002} & \res{0.56}{0.00} & \res{36.36}{0.02} & \res{0.50}{0.000} \\
			
			\midrule
			WRN & ASHA & \res{\,\,7.34}{0.76} & \res{0.91}{0.030} &\res{0.43}{0.01}  & \res{15.84}{0.52}& \res{0.41}{0.002} \\
			DARTS & GAEA & \res{17.67}{1.39} & \res{0.94}{0.020} & \res{0.34}{0.01}& \underline{\bf\res{12.51}{0.24}} &  \res{0.36}{0.020}\\ 
			
			\midrule 		
			AMBER & ENAS&n/a& n/a& \underline{\res{0.33}{0.02}}&\res{12.97}{0.07}& \underline{\res{0.32}{0.010}} \\ 	

			\midrule 				
			Expert & default & \res{8.73}{0.90} & \res{0.62}{0.004} & \textbf{\res{0.28}{0.00}} & \res{19.80}{0.00} &\textbf{\res{0.30}{0.024}} \\			
			
			\bottomrule
		\end{tabular}
		\begin{tablenotes}
			\item[$*$] did not fit on a single V100 GPU.
		\end{tablenotes}
	\end{threeparttable}
\end{table}

Having detailed our construction of NAS-Bench-360, in this section we will establish the experimental setup for our analyses in the following section, which demonstrates the usefulness of NAS-Bench-360 for evaluating NAS methods on diverse tasks. We first specify the NAS methods and baselines we compare, followed by the details of the experimental setup and intended use of the benchmark. Finally, we provide details of the precomputed NAS-Bench-201 search space for two representative diverse tasks from NAS-Bench-360: NinaPro and Darcy Flow.\looseness-1

\subsection{Baselines and Search Procedures}\label{sec:method}

Our initial experiments follow two practitioners with different resource settings: one with enough compute to tune a WRN (less-constrained) and another who can only train it once with the default hyperparameters (constrained). 
Given these two scenarios, we compare against NAS methods that each practitioner would be able to run.
In both cases, we focus on two well-known search paradigms:
cell-based NAS (using DARTS~\citep{liu2019darts}) and macro NAS (using DenseNAS~\citep{fang2020densenas}).
We further compare these approaches to two customized NAS methods: Auto-DeepLab \citep{liu2019auto} for 2D dense prediction and AMBER \citep{zhang2021automated} for 1D prediction, as well as general-purpose baselines: Perceiver IO \cite{jaegle2022perceiver} and XGBoost \cite{xgboost}. 
Additional details are provided in the Appendix. 


\begin{table}
	\centering
	\begin{threeparttable}
		\caption{Median rank and performance improvement over WRN across NAS-Bench-360.}
		\label{table-3}
		
		\footnotesize
		\begin{tabular}{lcccccccc}
			\toprule
			
			Search space   & WRN & DenseNAS & DenseNAS & WRN &  DARTS  & Auto-DL & AMBER \\ 
			Search algorithm & default & original & random & ASHA & GAEA & DARTS & ENAS  \\ 
			
			\midrule
			Median rank & 4.0&4.0 &4.0 & 3.5&1.5& 6.0$^\dagger$ & 1.0$^\dagger$ \\ 
			$\%$ better than WRN$^\ast$ & 0.0\%& 2.53\%& 0.0\% &0.1\%& 14.6\%& -75.3\%$^\dagger$ & 20.0\%$^\dagger$ \\ 
			
			\bottomrule
		\end{tabular}
		\begin{tablenotes}
			\item[$\ast$] relative improvement over the default (untuned) WRN baseline
			\item[$\dagger$] metric computed only on the subset of three tasks on which the method was evaluated
		\end{tablenotes}
	\end{threeparttable}
\end{table}

\begin{figure}[!t]
	\centering
	\includegraphics[width=1.0\linewidth]{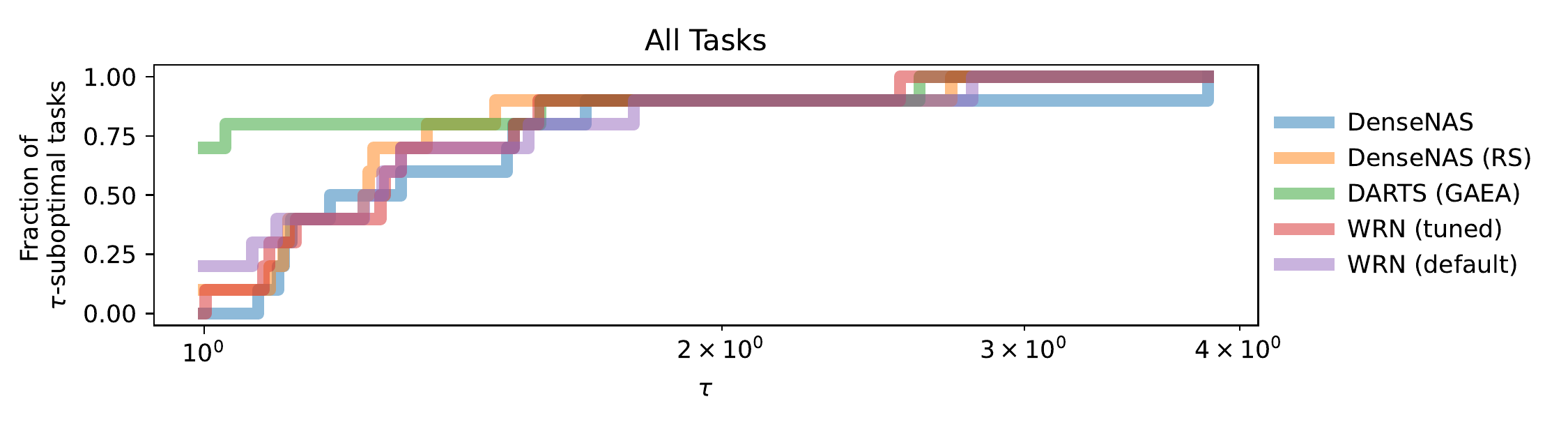}
	\\
	\vspace{-3mm}
	\includegraphics[width=1.0\linewidth]{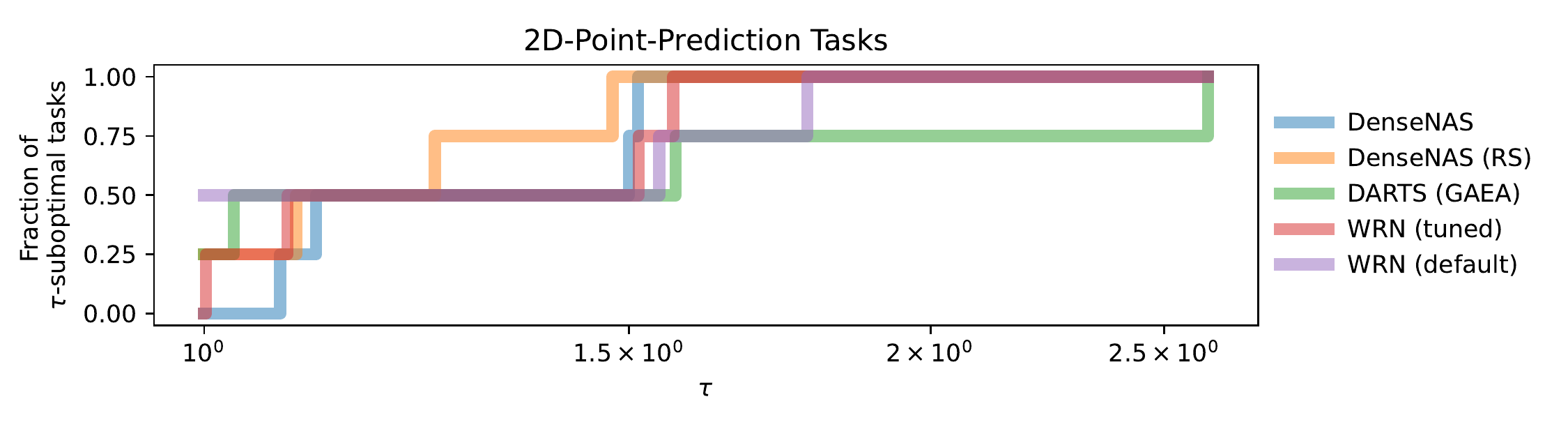}
	\vspace{-3mm}
	\caption{\label{fig:performance}
		In our investigation of modern methods on NAS-Bench-360, we find that methods like GAEA DARTS can be strong in aggregate, as shown in the performance profiles on all tasks (top), but worse on salient subsets such as 2D point tasks (bottom).
		The y-axis is the fraction of tasks on which error is within a factor $\tau$ of the optimal method, i.e. higher is better. 
	}
\end{figure}

\subsection{Experimental Setup}\label{sec:experiment}

Below we discuss the main reporting details of our empirical evaluation.


\begin{itemize}[leftmargin=*,topsep=-1pt,noitemsep]
	\item {\bf Hyperparameter tuning:} As detailed in the Appendix, we use the same hyperparameter ranges across all tasks to tune WRN. We use ASHA~\citep{li2018system} to search over these hyperparameters and give it a budget on each task that matches the total search and retraining budget of DARTS (GAEA).
	\item {\bf Aggregation metrics:} 
	To aggregate results across tasks, we use the median rank of each method and its performance improvement over WRN for direct comparison via a singe number, as demonstrated in Table~\ref{table-3}.
	However, since these metrics can be sensitive to small differences in performance, we also employ performance profiles \citep{dolan2002benchmarking} to mitigate that effect while still accounting for outliers.
	As described in Figure~\ref{fig:comparison}, these curves denote for each $\tau$ the fraction of tasks on which a method is no worse than a $\tau$-factor from the optimal. Concretely, we plot $ \rho_s(\tau) = \frac{1}{|\mathcal{P}|} \left| \left\{ p \in \mathcal{P}:  \frac{\text{error}_{p, s}}{ \min_{s \in \mathcal{S}} \text{error}_{p, s} } \leq \tau \right\} \right|$ given some method $s \in \mathcal{S}$ on tasks $\mathcal{P}$. 
	\item {\bf Software and hardware:} We adopt the free, open-source software \textit{Determined}\footnote{\url{https://github.com/determined-ai/determined}} for experiment management, hyperparameter tuning, and cloud deployment. All experiments are performed on a single p3.2xlarge instance with a 16GB NVIDIA V100 GPU. 
	While evaluation on NAS-Bench-360 indeed assumes access to at least a single V100 GPU, we reiterate that we provide the precomputed NAS-Bench-201 search space for two of our tasks in cases where GPU access is limited. 
	Costs in GPU-hours are in the appendix. 
\end{itemize}
	
\subsection{Precomputing NAS-Bench-201 on NinaPro and Darcy Flow}\label{sec:precompute}
The intended goal of NAS-Bench-360 is to evaluate the performance of \textit{NAS search method and search space pairs} on diverse tasks, which precludes the precomputation of all architectures in general due to the lack of a single fixed search space. 
A complete lack of precomputed architectures would be perhaps limiting for many NAS researchers, who rely on precomputed NAS benchmarks when developing new search methods. 
In an effort to address this potential limitation, we precompute all architectures in the NAS-Bench-201 \citep{dong2020nasbench201} search space on two representative tasks in NAS-Bench-360: NinaPro and Darcy Flow. We follow the same experimental procedure as in the original NAS-Bench-201 benchmark \citep{dong2020nasbench201} to generate the precompute results, except where they vary the number of models trained for each architecture between one and three, we fix the number of trials per architecture to one.
Note that NAS-Bench-201 already includes precompute for CIFAR-100, a dataset we include in NAS-Bench-360.\looseness-1

\section{Analysis} \label{analysis}

We conclude our presentation of NAS-Bench-360 with three sets of analyses. 
The first, a performance analysis of NAS methods and fixed baselines across diverse tasks,
reveals new insights about the capabilities and robustness of current NAS methods and demonstrates how our benchmark can enable critical next steps in NAS research.
In our second analysis, we evaluate claims from the NAS literature originally made using computer vision tasks, and show that they do not generalize to diverse tasks;
this demonstrates how NAS research can benefit from our contribution in the future.
Finally, we extend an existing analysis of zero-cost proxy methods on diverse tasks that already uses NAS-Bench-360 \cite{colin2022adeeperlook}.\looseness-1

\subsection{Performance across diverse tasks using NAS-Bench-360}

As discussed in Section~\ref{sec:experiment}, we start by considering two practitioners faced with a choice of spending their limited compute on a (possibly tuned) fixed-architecture CNN or trying to find a better architecture using NAS.
With this study, we investigate whether modern NAS methods perform well beyond the tasks for which they were designed.
\begin{enumerate}[leftmargin=*,topsep=-1pt,noitemsep]\setlength\itemsep{2pt}
	\item A surface-level analysis suggests that under light resource constraints, modern NAS in the form of DARTS (GAEA) is quite robust to a wide variety of tasks: Table~\ref{table-3} shows it is the highest-ranked domain-independent method and attains the most significant improvement over the fixed WRN baseline.
	The performance profile in Figure~\ref{fig:performance} (left) also seems favorable, although it is overtaken by tuned WRN at a higher $\tau$-suboptimality.
	However, a closer look at 2D point tasks in Figure~\ref{fig:performance} (right) reveals that DARTS is quite poor there, despite its design domain being image classification;
	in particular, it performs very poorly on NinaPro and FSD50K.
	Furthermore, on tasks where it performs well, it can still lag behind expert architectures;
	for example, on Darcy Flow, networks that use FNO \citep{li2021fno} or XD-operations \citep{roberts2021xd} do much better.
	Overall, our results suggest that this practitioner can apply NAS and expect to see some improvement, but also risks catastrophically poor performance (e.g. FSD50K) or not getting truly state-of-the-art results (e.g. Darcy Flow).
	\item Under stronger budget constraints, our experiments strongly suggest that a practitioner should simply apply the default WRN to their problem rather than undergo the additional complexity of using DenseNAS, as the latter attains little-to-no improvement over the former in Table~\ref{table-3} and has a usually-worse performance profiles Figure~\ref{fig:performance}.
	On the other hand, DenseNAS performs well on FSD50K---it outperforms all methods even while DARTS (GAEA) fails.
\end{enumerate}

These first experiments suggest that the modern NAS methods are not always robust to diverse tasks, especially under resource-constrained settings. 
We believe that NAS-Bench-360's main roles as a future benchmark include developing an understanding of the multi-domain performance of existing approaches and guiding research into better NAS methods.
While the latter is beyond the scope of this paper, our additional experiments demonstrate how NAS-Bench-360 facilitates the former. 

Notably, several of our results address the question of the relative importance of search space vs. search algorithm. 
For example, Table~\ref{table-3} shows that on DenseNAS, random search is nearly identical to the more sophisticated weight-sharing scheme of the original paper;
the two algorithms' performance profiles are also difficult to distinguish in Figure~\ref{fig:performance}.
Furthermore, AMBER---a 1D NAS method whose search space includes larger-kernel convolutions for handling such tasks---does better than GAEA even though it uses an older search algorithm (ENAS).
These both suggest that search space design, including the use of a wider variety of operations, may be at least as crucial for success as the search algorithm.
This point is reinforced by example tasks such as Darcy Flow, where architectures with more exotic operations substantially outperform our best results, as discussed earlier.

NAS-Bench-360 also reveals failure points of several methods, not just of general ones that usually perform quite well such as DARTS (GAEA) but also the objective-specific approach Auto-DL, which despite being designed for dense prediction tasks does poorly on all those considered here.
Understanding when and why these performance drops happen is critical to developing a more robust NAS that is useful not just on average but in more challenging settings.

\subsection{Do past NAS-Bench-201 analyses generalize to NAS-Bench-360?}

Existing NAS-benches have been widely used for analyses such as (1) comparing performances of different architectures across tasks, (2) quickly evaluating search methods, and (3) investigating design choices that impact performance.
In this section we show via the NAS-Bench-201 search space that the conclusions of past analyses cannot be assumed to hold on tasks beyond computer vision.\looseness-1

\subsubsection{Architecture transferability}

\begin{figure}[!t]
	\centering
	\includegraphics[width=0.9\linewidth]{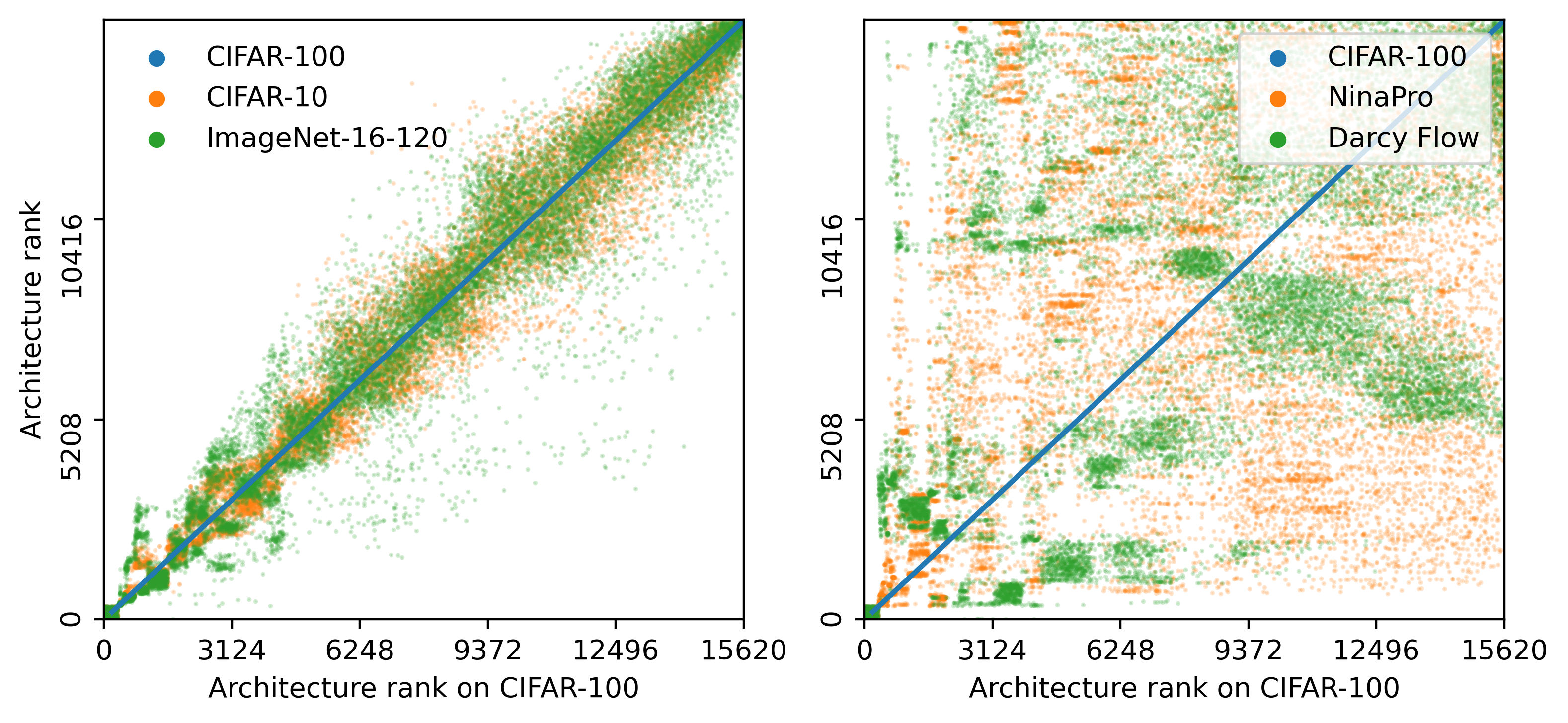}
	\caption{\label{fig:ranking}
		Architecture rankings between computer vision tasks correlate on NAS-Bench-201 \citep{dong2020nasbench201} (left, sorted by performance on CIFAR-100) but are uncorrelated between CIFAR-100 and two NAS-Bench-360 tasks, NinaPro and Darcy Flow (right).
	}
\end{figure}

We start by using the precomputed results outlined in Section~\ref{sec:precompute} to show in Figure~\ref{fig:ranking} the rank of each architecture across different datasets, indexed on the x-axis by its rank on CIFAR-100.
This reveals that while architecture rankings are highly correlated on image classification datasets---as pointed out by the authors of the original benchmark~\cite{dong2020nasbench201}---the rankings become uncorrelated when evaluated on a more diversified set of tasks. Therefore, NAS evaluations should be done across domains to verify true generalizability, and NAS-Bench-360 is especially useful for this purpose.\looseness-1


\subsubsection{Search algorithm performance}

Using the precomptued evaluations on two new datasets, we evaluate all ten typical NAS algorithms originally studied on NAS-Bench-201 \cite{dong2020nasbench201}. The results are shown in the Appendix. With similar wall clock time, the non-weight-sharing NAS algorithms that we evaluate: REINFORCE~\cite{williams1992simple}, random search (RS)~\cite{bergstra2012rs}, regularized evolution (REA)~\cite{real2019nas}, BOHB~\cite{falkner2018bohb}, and Hyperband~\cite{li2018hyperband} consistently perform well. Our results corroborate the strong performance of non-weight-sharing methods on this search space.

On the other hand, our experiments reveal some important differences for weight-sharing methods.
In particular, unlike in past experiments on the NAS-Bench-201 search space, DARTS does not always yield a network of all skip-connection on Darcy Flow, despite this behavior on image classification and NinaPro.
Instead, both first-order and second-order DARTS often pick convolution operations and sometimes achieve good performance, although still worse overall than the best non-weight-sharing methods.
These results together with the ranking demonstrate that evaluating methods and search spaces on vision tasks alone does not give a full picture of their capabilities and limitations, a problem alleviated by NAS-Bench-360.

\subsubsection{Operation redundancy} 

\begin{figure}[!t]
	\centering
	\includegraphics[width=0.2\linewidth]{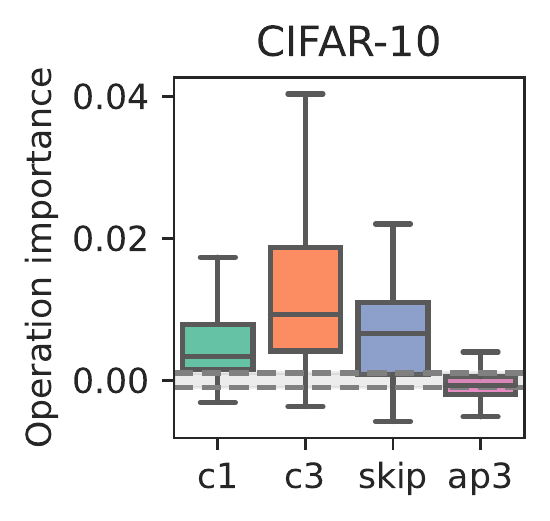}
	\includegraphics[width=0.19\linewidth]{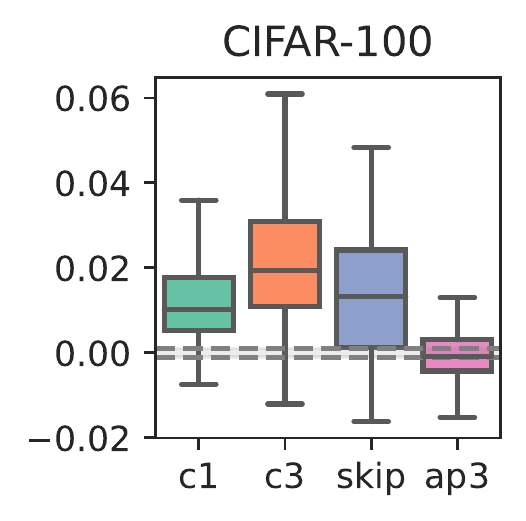}
	\includegraphics[width=0.19\linewidth]{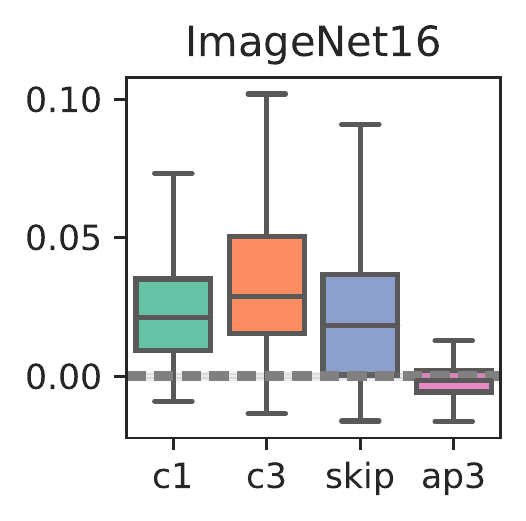}
	\includegraphics[width=0.19\linewidth]{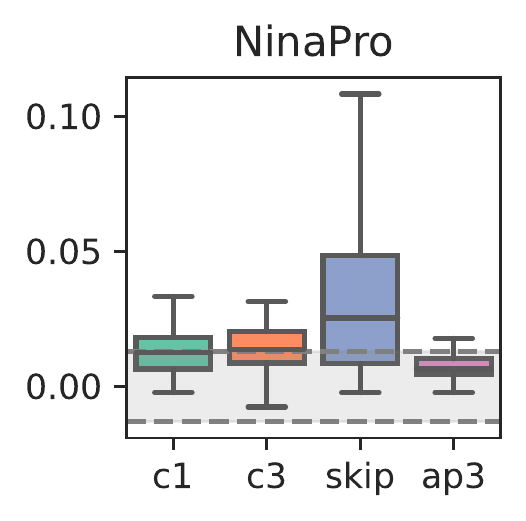}
	\includegraphics[width=0.19\linewidth]{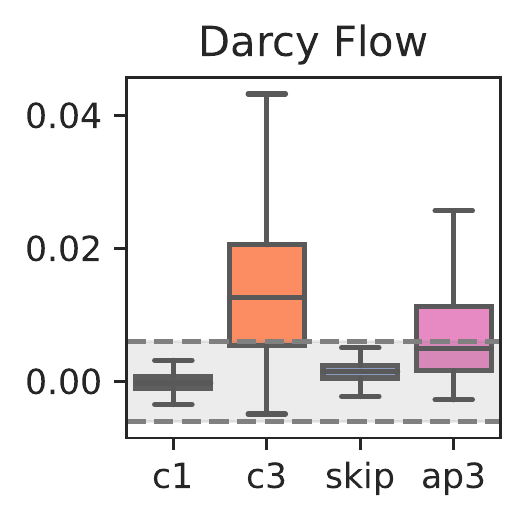}
	\caption{\label{fig:redundancy}
		Different operations are important for different tasks. While prior work \citep{wan2022on} shows that the operation importance distributions are stable across computer vision tasks---as shown by the high similarity of the three plots on the left---we find that they differ significantly for NinaPro and Darcy Flow.\looseness-1
	}
\end{figure}

Our final analysis using the NAS-Bench-201 search space is to investigate the conclusions of a more recent study on the redundancy of operations \cite{wan2022on}. 
We find that the operation redundancy phenomenon they outline is task-dependent and does not generalize to tasks beyond the three vision tasks---CIFAR-10, CIFAR-100, and ImageNet16-120---that they study.
To conduct our study we follow their procedure to obtain ``operation importance'' distributions for each operation in the NAS-Bench-201 search space for NinaPro and Darcy Flow; additionally, we reproduce their results on CIFAR-10, CIFAR-100, and ImageNet16-120. {\em Operation importance} measures the incremental effect of each operation choice in the NAS-Bench-201 search space---1x1 convolutions (c1), 3x3 convolutions (c3), skip connections (skip), and 3x3 average pooling (ap3)---on performance~\cite{wan2022on}. The original analysis found that the operation importance distributions are roughly similar across the original NAS-Bench-201 computer vision datasets, which we confirm and show in Figure~\ref{fig:redundancy}. However, we found that the operation importance distributions were drastically different for NinaPro and Darcy Flow, which we also show in Figure~\ref{fig:redundancy}. Not only are their distributions different from those of the computer vision tasks in the original analysis, but the operation importance distribution for NinaPro differs significantly from that of Darcy Flow. This tells us that \textit{different operations are more useful for different tasks}, and using NAS-Bench-360, we find that we cannot conclude that any of these operations are universally redundant or useful in a given search space across tasks. In other words, using NAS-Bench-360, we find that the original claim that ``existing search spaces contain a high degree of redundancy'' \cite{wan2022on} does not hold when considering diverse tasks beyond computer vision.\looseness-1

\subsection{Zero-cost proxies on diverse tasks} 

\begin{table}[!t]
	\centering
	\begin{threeparttable}
		\caption{Performance comparison of TE-NAS and GAEA using the DARTS search space on CIFAR-100, Spherical, NinaPro, and Darcy Flow. Lower is better for all metrics.}
		\label{table_4}
		
		\begin{tabular}{lcccccccc}
			\toprule
			
			& CIFAR-100 & Spherical & NinaPro & Darcy Flow \\
			TE-NAS & 24.32 & 56.87 & {\bf 9.71} & {\bf 0.012} \\
			GAEA & {\bf 24.02} & {\bf 48.23} & 17.67 & 0.026 \\
			
			\bottomrule
		\end{tabular}
	\end{threeparttable}
\end{table}

We conclude with an analysis of TE-NAS \cite{chen2021neural}, a zero-cost proxy inspired by neural tangent kernel (NTK) analysis, on four NAS-Bench-360 tasks.
Zero-cost proxies~\cite{mellor2021naswot,abdelfattah2021zcp} are the subject of a recent direction in NAS research that aims to construct quick-to-evaluate measures of architecture performance without doing any training.
Recently, \cite{colin2022adeeperlook} evaluated several zero-cost proxies on tasks from NAS-Bench-360 (Spherical, NinaPro, and Darcy Flow), as well as on TransNAS-Bench-101 \cite{duan2021transnas}. 
One major weakness of zero-cost proxies that they point out is that zero-cost proxies are not much more computationally efficient than weight-sharing methods, as the total compute cost is still dominated by the evaluation of the searched architecture~\cite{colin2022adeeperlook}. For example, this renders TE-NAS in the DARTS search space comparable to GAEA DARTS in terms of computational efficiency. 
The authors of \cite{colin2022adeeperlook} also point out that the performance of different zero-cost proxies vary considerably across diverse datasets, even subject to the same search space. Performance may be strong on some tasks, but weak on others.\looseness-1

To expand such study of zero-cost proxies, we look at one that \cite{colin2022adeeperlook} do not consider---TE-NAS---and evaluate its performance on the DARTS space using four NAS-Bench-360 tasks: CIFAR-100, Spherical, NinaPro, and Darcy Flow.
The results of this evaluation are shown in Table~\ref{table_4}. 
Unlike many other zero-cost-proxies~\citep{mellor2021naswot}, the fact that TE-NAS is constructed from a domain-agnostic NTK analysis rather than experiments makes it a potential candidate for good performance on diverse tasks.
However, Table~\ref{table_4} shows that performance does vary considerably across tasks, as observed for other proxies by \cite{colin2022adeeperlook}.
In-particular, TE-NAS performs okay on NinaPro and beats all methods in Table~\ref{table-2} on Darcy Flow---where its performance approaches that of the expert-designed FNO~\cite{li2021fno}---but does poorly on Spherical.
This evaluation adds evidence to existing scientific findings already enabled by NAS-Bench-360~\cite{colin2022adeeperlook} and provides additional evidence for the need to evaluate all NAS methods, including zero-cost proxies, on diverse tasks.\looseness-1

\section{Conclusion}\label{sec:conclusion}

NAS-Bench-360 is a new performance benchmark consisting of ten diverse tasks derived from various fields of research and practice.
It is designed for reproducible research on an academic budget that will guide the development of NAS methods and other automated approaches towards more robust performance across different domains.
In initial results, we have demonstrated both the need for such a benchmark and the utility of NAS-Bench-360 specifically for developing new search spaces and algorithms.
We also provide precompute architectures from the NAS-Bench-201 search space on two of the ten tasks. 
While the precomputed architectures on these two tasks are useful for analysis on their own, adding more precomputed search spaces and tasks is an area of further improvement.
We welcome researchers to use the NAS-Bench-360 tasks to develop new procedures for automating ML.

\section*{Acknowledgments}
We thank Maria-Florina Balcan for providing useful feedback. We also thank Hewlett Packard Enterprise for compute resources and the Determined AI open-source community for its support. This work was supported in part by DARPA FA875017C0141, the National Science Foundation grants IIS1705121, IIS1838017, IIS2046613, IIS-2112471, CCF2106707, the American Family Funding Initiative, the Wisconsin Alumni Research Foundation (WARF), an Amazon Web Services Award, a Facebook Faculty Research Award, funding from Booz Allen Hamilton Inc., a Block Center Grant, a Two Sigma Fellowship Award, and a Facebook PhD Fellowship Award. 
Any opinions, findings and conclusions or recommendations expressed in this material are those of the author(s) and do not necessarily reflect the views of any of these funding agencies.

\bibliography{_references,refs}

\begin{thebibliography}{10}

\bibitem{abdelfattah2021zcp}
Mohamed~S. Abdelfattah, Abhinav Mehrotra, Lukasz Dudziak, and Nicholas~D. Lane.
\newblock Zero-cost proxies for lightweight {NAS}.
\newblock In {\em Procedings of the 9th International Conference on Learning
  Representations}, 2021.

\bibitem{adhikari2020deepcon}
Badri Adhikari.
\newblock Deepcon: protein contact prediction using dilated convolutional
  neural networks with dropout.
\newblock {\em Bioinformatics}, 36(2):470--477, 2020.

\bibitem{adhikari2020fully}
Badri Adhikari.
\newblock A fully open-source framework for deep learning protein real-valued
  distances.
\newblock {\em Scientific reports}, 10(1):1--10, 2020.

\bibitem{atzori2012building}
Manfredo Atzori, Arjan Gijsberts, Simone Heynen, Anne-Gabrielle~Mittaz Hager,
  Olivier Deriaz, Patrick Van Der~Smagt, Claudio Castellini, Barbara Caputo,
  and Henning M{\"u}ller.
\newblock Building the ninapro database: A resource for the biorobotics
  community.
\newblock In {\em 2012 4th IEEE RAS \& EMBS International Conference on
  Biomedical Robotics and Biomechatronics (BioRob)}, pages 1258--1265. IEEE,
  2012.

\bibitem{bergstra2012rs}
James Bergstra and Yoshua Bengio.
\newblock Random search for hyper-parameter optimization.
\newblock {\em Journal of Machine Learning Research}, 13:281--305, 2012.

\bibitem{cai2019proxyless}
Han Cai, Ligeng Zhu, and Song Han.
\newblock Proxyless{NAS}: Direct neural architecture search on target task and
  hardware.
\newblock In {\em Proceedings of the 7th International Conference on Learning
  Representations}, 2019.

\bibitem{xgboost}
Tianqi Chen and Carlos Guestrin.
\newblock Xgboost: A scalable tree boosting system.
\newblock In {\em Proceedings of the 22nd ACM SIGKDD International Conference
  on Knowledge Discovery and Data Mining}, KDD '16, page 785–794, New York,
  NY, USA, 2016. Association for Computing Machinery.

\bibitem{chen2021neural}
Wuyang Chen, Xinyu Gong, and Zhangyang Wang.
\newblock Neural architecture search on imagenet in four {\{}gpu{\}} hours: A
  theoretically inspired perspective.
\newblock In {\em International Conference on Learning Representations}, 2021.

\bibitem{clifford2017af}
Gari~D Clifford, Chengyu Liu, Benjamin Moody, H~Lehman Li-wei, Ikaro Silva,
  Qiao Li, AE~Johnson, and Roger~G Mark.
\newblock Af classification from a short single lead ecg recording: The
  physionet/computing in cardiology challenge 2017.
\newblock In {\em 2017 Computing in Cardiology (CinC)}, pages 1--4. IEEE, 2017.

\bibitem{cohen2018spherical}
Taco~S. Cohen, Mario Geiger, Jonas Koehler, and Max Welling.
\newblock Spherical {CNN}s.
\newblock In {\em Proceedings of the 6th International Conference on Learning
  Representations}, 2018.

\bibitem{encode2004encode}
ENCODE~Project Consortium et~al.
\newblock The encode (encyclopedia of dna elements) project.
\newblock {\em Science}, 306(5696):636--640, 2004.

\bibitem{cote2019deep}
Ulysse C{\^o}t{\'e}-Allard, Cheikh~Latyr Fall, Alexandre Drouin, Alexandre
  Campeau-Lecours, Cl{\'e}ment Gosselin, Kyrre Glette, Fran{\c{c}}ois
  Laviolette, and Benoit Gosselin.
\newblock Deep learning for electromyographic hand gesture signal
  classification using transfer learning.
\newblock {\em IEEE Transactions on Neural Systems and Rehabilitation
  Engineering}, 27(4):760--771, 2019.

\bibitem{dempster2020rocket}
Angus Dempster, Fran{\c{c}}ois Petitjean, and Geoffrey~I Webb.
\newblock Rocket: exceptionally fast and accurate time series classification
  using random convolutional kernels.
\newblock {\em Data Mining and Knowledge Discovery}, 34(5):1454--1495, 2020.

\bibitem{dolan2002benchmarking}
Elizabeth~D Dolan and Jorge~J Mor{\'e}.
\newblock Benchmarking optimization software with performance profiles.
\newblock {\em Mathematical programming}, 91(2):201--213, 2002.

\bibitem{Dong2019OneShotNA}
Xuanyi Dong and Yezhou Yang.
\newblock One-shot neural architecture search via self-evaluated template
  network.
\newblock {\em 2019 IEEE/CVF International Conference on Computer Vision
  (ICCV)}, pages 3680--3689, 2019.

\bibitem{dong2019gdas}
Xuanyi Dong and Yi~Yang.
\newblock Searching for a robust neural architecture in four {GPU} hours.
\newblock In {\em Proceedings of the IEEE Conference on Computer Vision and
  Pattern Recognition}, 2019.

\bibitem{dong2020nasbench201}
Xuanyi Dong and Yi~Yang.
\newblock {NAS-Bench-201}: Extending the scope of reproducible neural
  architecture search.
\newblock In {\em Proceedings of the 8th International Conference on Learning
  Representations}, 2020.

\bibitem{duan2021transnas}
Yawen Duan, Xin Chen, Hang Xu, Zewei Chen, Xiaodan Liang, Tong Zhang, and
  Zhenguo Li.
\newblock {TransNAS-Bench-101}: Improving transferability and generalizability
  of cross-task neural architecture search.
\newblock In {\em Proceedings of the IEEE Conference on Computer Vision and
  Pattern Recognition}, 2021.

\bibitem{Duan2021TransNASBench101IT}
Yawen Duan, Xin Chen, Hang Xu, Zewei Chen, Li~Xiaodan, Tong Zhang, and Zhenguo
  Li.
\newblock Transnas-bench-101: Improving transferability and generalizability of
  cross-task neural architecture search.
\newblock {\em 2021 IEEE/CVF Conference on Computer Vision and Pattern
  Recognition (CVPR)}, pages 5247--5256, 2021.

\bibitem{elsken2019nas}
Thomas Elsken, Jan~Hendrik Metzen, and Frank Hutter.
\newblock Neural architecture search: A survey.
\newblock {\em Journal of Machine Learning Research}, 20(55):1--21, 2019.

\bibitem{falkner2018bohb}
Stefan Falkner, Aaron Klein, and Frank Hutter.
\newblock Bohb: Robust and efficient hyperparameter optimization at scale.
\newblock In {\em International Conference on Machine Learning}, pages
  1437--1446. PMLR, 2018.

\bibitem{fang2020densenas}
Jiemin Fang, Yuzhu Sun, Qian Zhang, Yuan Li, Wenyu Liu, and Xinggang Wang.
\newblock Densely connected search space for more flexible neural architecture
  search.
\newblock In {\em Proceedings of the IEEE Conference on Computer Vision and
  Pattern Recognition}, 2020.

\bibitem{fonseca2020fsd50k}
Eduardo Fonseca, Xavier Favory, Jordi Pons, Frederic Font, and Xavier Serra.
\newblock Fsd50k: an open dataset of human-labeled sound events.
\newblock {\em arXiv preprint arXiv:2010.00475}, 2020.

\bibitem{fonseca2017freesound}
Eduardo Fonseca, Jordi Pons~Puig, Xavier Favory, Frederic Font~Corbera, Dmitry
  Bogdanov, Andres Ferraro, Sergio Oramas, Alastair Porter, and Xavier Serra.
\newblock Freesound datasets: a platform for the creation of open audio
  datasets.
\newblock In {\em Hu X, Cunningham SJ, Turnbull D, Duan Z, editors. Proceedings
  of the 18th ISMIR Conference; 2017 oct 23-27; Suzhou, China.[Canada]:
  International Society for Music Information Retrieval; 2017. p. 486-93.}
  International Society for Music Information Retrieval (ISMIR), 2017.

\bibitem{garofolo1993timit}
John~S Garofolo.
\newblock Timit acoustic phonetic continuous speech corpus.
\newblock {\em Linguistic Data Consortium, 1993}, 1993.

\bibitem{gemmeke2017audio}
Jort~F Gemmeke, Daniel~PW Ellis, Dylan Freedman, Aren Jansen, Wade Lawrence,
  R~Channing Moore, Manoj Plakal, and Marvin Ritter.
\newblock Audio set: An ontology and human-labeled dataset for audio events.
\newblock In {\em 2017 IEEE International Conference on Acoustics, Speech and
  Signal Processing (ICASSP)}, pages 776--780. IEEE, 2017.

\bibitem{he2016resnet}
Kaiming He, Xiangyu Zhang, Shaoqing Ren, and Jian Sun.
\newblock Deep residual learning for image recognition.
\newblock In {\em Proceedings of the IEEE Conference on Computer Vision and
  Pattern Recognition}, 2016.

\bibitem{hong2020holmes}
Shenda Hong, Yanbo Xu, Alind Khare, Satria Priambada, Kevin Maher, Alaa
  Aljiffry, Jimeng Sun, and Alexey Tumanov.
\newblock Holmes: health online model ensemble serving for deep learning models
  in intensive care units.
\newblock In {\em Proceedings of the 26th ACM SIGKDD International Conference
  on Knowledge Discovery \& Data Mining}, pages 1614--1624, 2020.

\bibitem{huang2017densely}
Gao Huang, Zhuang Liu, Laurens Van Der~Maaten, and Kilian~Q Weinberger.
\newblock Densely connected convolutional networks.
\newblock In {\em Proceedings of the IEEE conference on computer vision and
  pattern recognition}, pages 4700--4708, 2017.

\bibitem{jaegle2022perceiver}
Andrew Jaegle, Sebastian Borgeaud, Jean-Baptiste Alayrac, Carl Doersch, Catalin
  Ionescu, David Ding, Skanda Koppula, Daniel Zoran, Andrew Brock, Evan
  Shelhamer, Olivier~J Henaff, Matthew Botvinick, Andrew Zisserman, Oriol
  Vinyals, and Joao Carreira.
\newblock Perceiver {IO}: A general architecture for structured inputs \&
  outputs.
\newblock In {\em International Conference on Learning Representations}, 2022.

\bibitem{josephs2020semg}
David Josephs, Carson Drake, Andy Heroy, and John Santerre.
\newblock semg gesture recognition with a simple model of attention.
\newblock In {\em Machine Learning for Health}, pages 126--138. PMLR, 2020.

\bibitem{alphafold2}
John Jumper, Richard Evans, Alexander Pritzel, Tim Green, Michael Figurnov,
  Kathryn Tunyasuvunakool, Olaf Ronneberger, Russ Bates, Augustin {\v
  Z}{\'\i}dek, Alex Bridgland, Clemens Meyer, Simon A~A Kohl, Anna Potapenko,
  Andrew~J Ballard, Andrew Cowie, Bernardino Romera-Paredes, Stanislav Nikolov,
  Rishub Jain, Jonas Adler, Trevor Back, Stig Petersen, David Reiman, Martin
  Steinegger, Michalina Pacholska, David Silver, Oriol Vinyals, Andrew~W
  Senior, Koray Kavukcuoglu, Pushmeet Kohli, and Demis Hassabis.
\newblock High accuracy protein structure prediction using deep learning.
\newblock {\em In Fourteenth Critical Assessment of Techniques for Protein
  Structure Prediction (Abstract Book)}, 2020.

\bibitem{klyuchnikov2020nlp}
Nikita Klyuchnikov, Ilya Trofimov, Ekaterina Artemova, Mikhail Salnikov, Maxim
  Fedorov, and Evgeny Burnaev.
\newblock {NAS-Bench-NLP}: Neural architecture search benchmark for natural
  language processing.
\newblock arXiv, 2020.

\bibitem{krizhevsky2009cifar}
Alex Krizhevksy.
\newblock Learning multiple layers of features from tiny images.
\newblock Technical report, 2009.

\bibitem{li2018hyperband}
Liam Li, Kevin Jamieson, Giulia DeSalvo, Afshin Rostamizadeh, and Ameet
  Talwalkar.
\newblock Hyperband: A novel bandit-based approach to hyperparameter
  optimization.
\newblock {\em Journal of Machine Learning Research}, 18(185):1--52, 2018.

\bibitem{li2018system}
Liam Li, Kevin Jamieson, Afshin Rostamizadeh, Ekaterina Gonina, Moritz Hardt,
  Benjamin Recht, and Ameet Talwalkar.
\newblock A system for massively parallel hyperparameter tuning.
\newblock {\em arXiv preprint arXiv:1810.05934}, 2018.

\bibitem{li2021gaea}
Liam Li, Mikhail Khodak, Maria-Florina Balcan, and Ameet Talwalkar.
\newblock Geometry-aware gradient algorithms for neural architecture search.
\newblock In {\em Proceedings of the 9th International Conference on Learning
  Representations}, 2021.

\bibitem{li2019rsws}
Liam Li and Ameet Talwalkar.
\newblock Random search and reproducibility for neural architecture search.
\newblock In {\em Proceedings of the Conference on Uncertainty in Artificial
  Intelligence}, 2019.

\bibitem{li2017hyperband}
Lisha Li, Kevin~G Jamieson, Giulia DeSalvo, Afshin Rostamizadeh, and Ameet
  Talwalkar.
\newblock Hyperband: Bandit-based configuration evaluation for hyperparameter
  optimization.
\newblock In {\em ICLR (Poster)}, 2017.

\bibitem{li2021fno}
Zongyi Li, Nikola~Borislavov Kovachki, Kamyar Azizzadenesheli, Burigede Liu,
  Kaushik Bhattacharya, Andrew Stuart, and Anima Anandkumar.
\newblock Fourier neural operator for parametric partial differential
  equations.
\newblock In {\em Proceedings of the 9th International Conference on Learning
  Representations}, 2021.

\bibitem{liu2019auto}
Chenxi Liu, Liang-Chieh Chen, Florian Schroff, Hartwig Adam, Wei Hua, Alan~L
  Yuille, and Li~Fei-Fei.
\newblock Auto-deeplab: Hierarchical neural architecture search for semantic
  image segmentation.
\newblock In {\em Proceedings of the IEEE/CVF Conference on Computer Vision and
  Pattern Recognition}, pages 82--92, 2019.

\bibitem{liu2019darts}
Hanxiao Liu, Karen Simonyan, and Yiming Yang.
\newblock {DARTS}: Differentiable architecture search.
\newblock In {\em Proceedings of the 7th International Conference on Learning
  Representations}, 2019.

\bibitem{mehrotra2021asr}
Abhinav Mehrotra, Alberto Gil, C.~P. Ramos, Sourav Bhattacharya, {\L}ukasz
  Dudziak, Ravichander Vipperla, Thomas Chau, Samin Ishtiaq, Mohamed~S.
  Abdelfattah, and Nicholas~D. Lane.
\newblock {NAS-Bench-ASR}: Reproducible neural architecture search for speech
  recognition.
\newblock In {\em Proceedings of the 8th International Conference on Learning
  Representations}, 2021.

\bibitem{mellor2021naswot}
Joseph Mellor, Jack Turner, Amos Storkey, and Elliot~J. Crowley.
\newblock Neural architecture search without training.
\newblock In {\em Proceedings of the 38th International Conference on Machine
  Learning}, 2021.

\bibitem{petitjean2012satellite}
Fran{\c{c}}ois Petitjean, Jordi Inglada, and Pierre Gan{\c{c}}arski.
\newblock Satellite image time series analysis under time warping.
\newblock {\em IEEE transactions on geoscience and remote sensing},
  50(8):3081--3095, 2012.

\bibitem{pham2018enas}
Hieu Pham, Melody~Y. Guan, Barret Zoph, Quoc~V. Le, and Jeff Dean.
\newblock Efficient neural architecture search via parameter sharing.
\newblock In {\em Proceedings of the 35th International Conference on Machine
  Learning}, 2018.

\bibitem{real2019nas}
Esteban Real, Alok Aggarwal, Yanping Huang, and Quoc~V. Le.
\newblock Regularized evolution for image classifier architecture search.
\newblock In {\em Proceedings of the 33rd AAAI Conference on Artificial
  Intelligence}, 2019.

\bibitem{real2020automlzero}
Esteban Real, Chen Liang, David~R. So, and Quoc~V. Le.
\newblock Auto{ML}-{Z}ero: Evolving machine learning algorithms from scratch.
\newblock In {\em Proceedings of the 37th International Conference on Machine
  Learning}, 2020.

\bibitem{roberts2021xd}
Nicholas Roberts, Mikhail Khodak, Tri Dao, Liam Li, Chris R\'e, and Ameet
  Talwalkar.
\newblock Rethinking neural operations for diverse tasks.
\newblock arXiv, 2021.

\bibitem{triantafillou2020metadataset}
Eleni Triantafillou, Tyler Zhu, Vincent Dumoulin, Pascal Lamblin, Utku Evci,
  Kelvin Xu, Ross Goroshin, Carles Gelada, Kevin Swersky, Pierre-Antoine
  Manzagol, and Hugo Larochelle.
\newblock Meta-dataset: A dataset of datasets for learning to learn from few
  examples.
\newblock In {\em Proceedings of the 8th International Conference on Learning
  Representations}, 2020.

\bibitem{wan2022on}
Xingchen Wan, Binxin Ru, Pedro~M Esperan{\c{c}}a, and Zhenguo Li.
\newblock On redundancy and diversity in cell-based neural architecture search.
\newblock In {\em International Conference on Learning Representations}, 2022.

\bibitem{colin2022adeeperlook}
Colin White, Mikhail Khodak, Renbo Tu, Shital Shah, Sébastien Bubeck, and
  Debadeepta Dey.
\newblock A deeper look at zero-cost proxies for lightweight nas.
\newblock In {\em ICLR Blog Track}, 2022.
\newblock https://iclr-blog-track.github.io/2022/03/25/zero-cost-proxies/.

\bibitem{white2021bananas}
Colin White, Willie Neiswanger, and Yash Savani.
\newblock {BANANAS}: Bayesian optimization with neural architectures for neural
  architecture search.
\newblock In {\em Proceedings of the 35th AAAI Conference on Artificial
  Intelligence}, 2021.

\bibitem{williams1992simple}
Ronald~J Williams.
\newblock Simple statistical gradient-following algorithms for connectionist
  reinforcement learning.
\newblock {\em Machine learning}, 8(3):229--256, 1992.

\bibitem{xu2020pcdarts}
Yuhui Xu, Lingxi Xie, Xiaopeng Zhang, Xin Chen, Guo-Jun Qi, Qi~Tian, and
  Hongkai Xiong.
\newblock {PC-DARTS}: Partial channel connections for memory-efficient
  architecture search.
\newblock In {\em Proceedings of the 8th International Conference on Learning
  Representations}, 2020.

\bibitem{ying2019nasbench101}
Chris Ying, Aaron Klein, Eric Christiansen, Esteban Real, Kevin Murphy, and
  Frank Hutter.
\newblock {NAS-Bench-101}: Towards reproducible neural architecture search.
\newblock In {\em Proceedings of the 36th International Conference on Machine
  Learning}, 2019.

\bibitem{zagoruyko2016wideresnet}
Sergey Zagoruyko and Nikos Komodakis.
\newblock Wide residual networks.
\newblock In {\em Proceedings of the British Machine Vision Conference}, 2016.

\bibitem{zela2020nasbench1shot1}
Arber Zela, Julien Siems, and Frank Hutter.
\newblock {NAS-Bench-1Shot1}: Benchmarking and dissecting one-shot neural
  architecture search.
\newblock In {\em Proceedings of the 8th International Conference on Learning
  Representations}, 2020.

\bibitem{zhang2020deepcr}
Keming Zhang and Joshua~S Bloom.
\newblock deepcr: Cosmic ray rejection with deep learning.
\newblock {\em The Astrophysical Journal}, 889(1):24, 2020.

\bibitem{zhang2021ambient}
Zijun Zhang, Evan~M Cofer, and Olga~G Troyanskaya.
\newblock Ambient: accelerated convolutional neural network architecture search
  for regulatory genomics.
\newblock {\em bioRxiv}, 2021.

\bibitem{zhang2021automated}
Zijun Zhang, Christopher~Y Park, Chandra~L Theesfeld, and Olga~G Troyanskaya.
\newblock An automated framework for efficiently designing deep convolutional
  neural networks in genomics.
\newblock {\em Nature Machine Intelligence}, 3(5):392--400, 2021.

\bibitem{zhou2015predicting}
Jian Zhou and Olga~G Troyanskaya.
\newblock Predicting effects of noncoding variants with deep learning--based
  sequence model.
\newblock {\em Nature methods}, 12(10):931--934, 2015.

\bibitem{zoph2018nas}
Barret Zoph, Vijay Vasudevan, Jonathon Shlens, and Quoc~V. Le.
\newblock Learning transferable architectures for scalable image recognition.
\newblock In {\em Proceedings of the IEEE Conference on Computer Vision and
  Pattern Recognition}, 2018.

\end{thebibliography}
\bibliographystyle{plainnat}

\section*{Checklist}


\begin{enumerate}

\item For all authors...
\begin{enumerate}
  \item Do the main claims made in the abstract and introduction accurately reflect the paper's contributions and scope?
    \answerYes{}
  \item Did you describe the limitations of your work?
    \answerYes{See Section~\ref{sec:conclusion}}
  \item Did you discuss any potential negative societal impacts of your work?
    \answerYes{See Appendix.}
  \item Have you read the ethics review guidelines and ensured that your paper conforms to them?
    \answerYes{}
\end{enumerate}

\item If you are including theoretical results...
\begin{enumerate}
  \item Did you state the full set of assumptions of all theoretical results?
    \answerNA{We do not include any theoretical results.}
	\item Did you include complete proofs of all theoretical results?
    \answerNA{We do not include any theoretical results.}
\end{enumerate}

\item If you ran experiments (e.g. for benchmarks)...
\begin{enumerate}
  \item Did you include the code, data, and instructions needed to reproduce the main experimental results (either in the supplemental material or as a URL)?
    \answerYes{Instructions are described in the paper; code and data are available on our dedicated website.}
  \item Did you specify all the training details (e.g., data splits, hyperparameters, how they were chosen)?
    \answerYes{In both Section~\ref{sec:tasks} and in the Appendix.}
	\item Did you report error bars (e.g., with respect to the random seed after running experiments multiple times)?
    \answerYes{See Table~\ref{table-2}}
	\item Did you include the total amount of compute and the type of resources used (e.g., type of GPUs, internal cluster, or cloud provider)?
    \answerYes{See Appendix.}
\end{enumerate}

\item If you are using existing assets (e.g., code, data, models) or curating/releasing new assets...
\begin{enumerate}
  \item If your work uses existing assets, did you cite the creators?
    \answerYes{}
  \item Did you mention the license of the assets?
    \answerYes{See Appendix.}
  \item Did you include any new assets either in the supplemental material or as a URL?
    \answerYes{New assets are on our website.}
  \item Did you discuss whether and how consent was obtained from people whose data you're using/curating?
    \answerYes{See Appendix.}
  \item Did you discuss whether the data you are using/curating contains personally identifiable information or offensive content?
    \answerYes{See Appendix.}
\end{enumerate}

\item If you used crowdsourcing or conducted research with human subjects...
\begin{enumerate}
  \item Did you include the full text of instructions given to participants and screenshots, if applicable?
    \answerNA{We do not crowdsource any data or conduct research with human subjects.}
  \item Did you describe any potential participant risks, with links to Institutional Review Board (IRB) approvals, if applicable?
    \answerNA{We do not crowdsource any data or conduct research with human subjects.}
  \item Did you include the estimated hourly wage paid to participants and the total amount spent on participant compensation?
    \answerNA{We do not crowdsource any data or conduct research with human subjects.}
\end{enumerate}

\end{enumerate}

\appendix
\section{Tabular Benchmark Results}
The precomputed evaluation on the NinaPro and Darcy Flow tasks significantly reduce runtimes of NAS algorithms. We evaluate the 10 search methods from NAS-Bench-201 \cite{dong2020nasbench201}, which comprises 6 non-weight sharing methods and 4 weight-sharing methods. The time budget of non-weight-sharing algorithms are set to match that of DARTS-V1. The non-weight-sharing methods are averaged over 500 runs, and the weight-sharing methods are averaged over 3 runs. 

\begin{table}
	\caption{Results of 12 NAS algorithms across three datasets of NAS-Bench-360. We report the mean and standard deviation. All metrics are errors such that lower is better. The best methods in the non-WS and WS groups are bolded.}
	\label{table-8}
	\centering
	\begin{tabular}{lccc}
		\toprule
		
		Algorithm     &  NinaPro  &  Darcy Flow &  CIFAR-100 \\ 
		\midrule
		REINFORCE~\cite{williams1992simple} & \textbf{\,\,\,8.07$\pm$0.73} & 0.0247$\pm$0.006& 28.14$\pm$0.89 \\
		RS~\cite{bergstra2012rs} & \,\,\,8.09$\pm$0.71 & 0.0252$\pm$0.006& 28.45$\pm$0.97\\
		REA~\cite{real2019nas} & \,\,\,8.15$\pm$0.85 & 0.0244$\pm$0.006& 27.77$\pm$0.84\\ 
		BOHB~\cite{falkner2018bohb} & \,\,\,8.17$\pm$0.57 &0.0194$\pm$0.002& 28.00$\pm$0.86\\ 
	         Hyperband~\cite{li2018hyperband} & \,\,\,8.16$\pm$0.57 &\textbf{0.0191$\pm$0.002}& \textbf{27.66$\pm$0.67}\\ 
		\midrule 
		DARTS-V1~\cite{liu2019darts}  & 22.06$\pm$2.00 & 0.178$\pm$0.107& 38.74$\pm$4.43\\ 
		DARTS-V2~\cite{liu2019darts}  &22.06$\pm$2.00&0.150$\pm$0.093&39.51$\pm$4.95\\
		RSWS~\cite{li2019rsws} &\,\,\,9.82$\pm$1.49&0.221$\pm$0.045&31.74$\pm$0.96 \\
		GDAS~\cite{dong2019gdas} &17.61$\pm$6.39&0.180$\pm$0.103&31.87$\pm$2.50 \\
		SETN~\cite{Dong2019OneShotNA} &14.56$\pm$7.30&0.253$\pm$0.000&30.64$\pm$1.72 \\
		ENAS~\cite{pham2018enas} &11.56$\pm$1.12&0.253$\pm$0.000&29.33$\pm$0.62 \\
		GAEA (ERM)~\cite{li2021gaea}  &\textbf{\,\,\,7.69$\pm$0.19}&\textbf{0.026$\pm$0.003}& \textbf{26.73$\pm$0.18} \\
		\bottomrule
		\multicolumn{4}{l}{\small *We did not include the zero operation in GAEA(ERM), as in the original paper. } \\

	\end{tabular}
\end{table}

\section{Dataset Descriptions}

\paragraph{CIFAR-100: Standard Image Classification}

As a starting point of comparison to existing benchmarks, we include the {\bf CIFAR-100} task \citep{krizhevsky2009cifar}, which contains RGB images from natural settings to be classified into 100 fine-grained categories. 
CIFAR-100 is preferred over CIFAR-10 because it is more challenging and suffers less from over-fitting in previous research. 

\paragraph{Spherical: Classifying Spherically Projected CIFAR-100 Images} 

To test NAS methods applied to natural-image-like data, we consider the task of classifying spherical projections of the CIFAR-100 images, which we call {\bf Spherical}.
In addition to scientific interest, spherical image data is also present in various applications, such as omnidirectional vision in robotics and weather modeling in meteorology, as sensors usually produce distorted image signals in real-life settings. 
To create Spherical CIFAR, we project the planar signals of the CIFAR images to the northern hemisphere and add a random rotation to produce spherical signals for each channel following the procedure specified in \cite{cohen2018spherical}. The resulting images are 60$\times$60 pixels with RGB channels.

\paragraph{NinaPro: Classifying Electromyography Signals}

{\bf NinaPro} moves away from the image domain to classify hand gestures indicated by electromyography signals.
For this, we use a subset of the NinaPro DB5 dataset \citep{atzori2012building} in which two Myo armbands collect EMG signals from 10 test individuals who hold 18 different hand gestures to be classified.
These armbands leverage data from muscle movement, which is collected using electrodes in the form of wave signals.
Each wave signal is then sampled using a wavelength and frequency prescribed in \cite{cote2019deep} to produce 2D signals.

\paragraph{FSD50K:  Labeling Sound Events}
{\bf FSD50K} \citep{fonseca2020fsd50k} is derived from the larger Freesound dataset \citep{fonseca2017freesound} of Youtube videos with 51,000 clips totaling more than 100 hours of sound. These clips are manually labeled and equally distributed in 200 classes from the AudioSet ontology \citep{gemmeke2017audio}. Each clip could receive multiple labels. Unlike TIMIT~\citep{garofolo1993timit}, FSD50K does not focus exclusively on sounds of spoken language but includes sound events from physical sources and production mechanisms. The mean average precision (mAP) is used to evaluate classification results.

\paragraph{Darcy Flow: Solving Partial Differential Equations (PDEs)} \label{sssec:pde}

Our first regression task, {\bf Darcy Flow}, focuses on learning a map from the initial conditions of a PDE to the solution at a later timestep.
This application aims to replace traditional solvers with learned neural networks, which can output a result in a single forward pass.
The input is a 2d grid specifying the initial conditions of a fluid, and the output is a 2d grid specifying the fluid state at a later time, with the ground truth being the result computed by a traditional solver.  
This we use the same Darcy Flow dataset that was used in \cite{li2021fno}.  
We report the mean square error (MSE or $\ell_2$). 

\paragraph{PSICOV: Protein Distance Prediction}

{\bf PSICOV} studies the use of neural networks in the protein folding prediction pipeline, which has recently received significant attention to the success of methods like AlphaFold \citep{alphafold2}.
While the dataset and method they use are too large-scale for our purposes, we consider a smaller set of protein structures to tackle the specific problem of inter-residual distance predictions outlined in \cite{adhikari2020fully}. 
 2D large-scale features are extracted from protein sequences, resulting in input feature maps with a massive number of channels. Correspondingly, the labels are pairwise-distance matrices with the same spatial dimension. The evaluation metric is mean absolute error (MAE or $\ell_1$) computed on distances below 8 \text{\normalfont\AA}, referred to as MAE$_8$.

\paragraph{Cosmic: Identifying Cosmic Ray Contamination}

Images from space-based facilities are prone to corruption by charged particles collectively referred to as "cosmic rays." Cosmic rays on images should be identified and masked before the images are used for further analysis \citep{zhang2020deepcr}. The \textbf{Cosmic} task uses imaging data of local resolved galaxies collected from the Hubble Space Telescope. Inputs and outputs are same-size 2D matrices, with the output predicting whether each pixel in the input is an artifact of cosmic rays. We report the false-negative rate (FNR) of identification results. 

\paragraph{ECG: Detecting Heart Disease}

Electrocardiograms are frequently used in medicine to diagnose sinus rhythm irregularities. The \textbf{ECG} task is based on the 2017 PhysioNet Challenge \citep{clifford2017af}, with 9 to 60-second ECG recordings sampled at 300 Hz and labeled using four classes: normal, disease, other, or noisy rhythms. Recordings are processed using a fixed sliding window of 1,000 ms and stride of 500 ms. 
We report the F1-score according to the challenge's guidelines.

\paragraph{Satellite: Satellite Image Time Series Analysis}

Satellite image time series (SITS) are becoming more widely available in earth monitoring applications. Our dataset comes from Formosat-2 satellite images acquired over Toulouse, France \citep{petitjean2012satellite}. Available in multiple channels, SITS track the land cover changes over several years as each pixel in the image represents a geographical region. The goal of the \textbf{Satellite} task is to generate land cover maps for geo-surveying. Specifically, a series of pixels in a given color channel which constitutes a time series to be classified into 
24
land cover types. 

\paragraph{DeepSEA: Predicting Functional Effects From Genetic Sequences}

Predicting chromatin effects of genetic sequence alterations is a significant challenge in the field to understand genetic diseases. \textbf{DeepSEA} \citep{zhou2015predicting}, provides a compendium of genomic profiles from the Encyclopedia of DNA Elements (ENCODE) project \citep{encode2004encode} to train a predictive model estimating the behavior of chromatin proteins, divided into 919 categories. Due to computation constraints, we subsample 36 of these categories as per \cite{zhang2021ambient} and further take 5\% of the training data for prediction. We report the area under the receiver operating characteristic (AUROC) following the previous work.

\section{Baselines}

\paragraph{Wide ResNet With Hyperparameter Tuning} 
Architectures based on ResNet \cite{he2016resnet} are a common first choice by practitioners faced with a new domain \citep{fonseca2020fsd50k,adhikari2020fully};
it is thus a natural source of fixed-architecture baselines for our study.
We use the Wide ResNet variant \citep{zagoruyko2016wideresnet} with 16 layers and a widen factor of 4, and apply its original training routine directly for the constrained practitioner.
For the other practitioner, we wrap the training procedure with a hyperparameter tuner, ASHA \citep{li2018system}, an asynchronous version of Hyperband \citep{li2017hyperband}.  Given a range for each hyperparameter, ASHA uniformly samples configurations and uses brackets of elimination: at each round, each configuration is trained for some epochs, before the algorithm selects the best-performing portion based on validation metrics. This procedure is useful for finding suitable hyperparameters in an easy-to-use, automated fashion. 

\paragraph{Cell-Based Search Using DARTS}
The first NAS paradigm we consider is cell-based NAS. These methods first search for a genotype, a cell containing neural operations. During evaluation, an architecture is constructed by replicating the searched cell and stacking them together. 
The most popular search space for this approach is DARTS \citep{liu2019darts}, which assigns one of eight operations to six edges in two types of cells: ``normal'' cells preserve the shape of the input while ``reduction'' cells downsample it.
For dense tasks, we do not use the reduction cell to prevent introducing a bottleneck. For 1D tasks, all convolutions in the cell are converted from 2D to 1D.
Finally, to adhere to standard ML practices, we do {\em not} adapt the standard DARTS pipeline from the original paper.
As this search space has been heavily studied, we use as a search routine a recent approach, GAEA PC-DARTS (GAEA), that achieves some of the best-known results on CIFAR-10 and ImageNet \citep{li2021gaea}. This NAS approach, due to its heavy retraining routine, is compared to the tuned WRN baseline of the less-resource-constrained practitioner.

\paragraph{Macro NAS Using DenseNAS}

The second NAS paradigm we consider is macro NAS. Instead of building from a fixed cell, it requires the specification of a supernet with different inter-connected blocks. These blocks and connections are then pruned to construct an architecture. For our benchmark, we choose a recent search space, DenseNAS \citep{fang2020densenas}, which has near state-of-the-art results on ImageNet. 
DenseNAS searches for architectures with densely-connected, customizable routing blocks to emulate DenseNet \citep{huang2017densely}. In our experiments, we use the ResNet-based search space, DenseNAS-R1, which contains all neural operations of the WRN backbone.  For 2D tasks, we adapt two super networks from the one used for ImageNet as inputs to the search algorithm. The super network for dense tasks maintains the same spatial dimensions without downsampling to avoid bottlenecks, and we lower the learning rate for evaluating architectures to prevent divergence. For transferring to 1D tasks, all network operations are switched from 2D to 1D. Other training and evaluation procedures are identical to those in the original paper and uniform across all tasks. DenseNAS is quick to search and evaluate, making it comparable to the fixed WRN baseline. 

We apply another search method to the fixed DenseNAS space to study the relative importance of search algorithms. Random search is implemented through randomly sampling architectures from the DenseNAS space and validating them after a brief training period of 10 epochs before evaluating the best performer. To ensure fairness of comparison, we allot equal GPU hours to random search and regular DenseNAS search, additionally applying a soft constraint that random architecture model sizes should not surpass DenseNAS searched architecture sizes significantly. 

\paragraph{Domain-Specific NAS Baselines: Auto-DL and AMBER}
To study the importance of search spaces, we further handpick two domain-specific NAS approaches applicable only to a subset of the tasks. Using an encoder-decoder architecture, Auto-DeepLab (Auto-DL) \citep{liu2019auto} is designed for dense prediction, e.g., semantic segmentation. While the decoder is fixed, Auto-DL searches for an encoder via first-order DARTS. We evaluate Auto-DL on Darcy Flow, PSICOV, and Cosmic tasks.

AMBER \citep{zhang2021automated} aims to automate neural network design for 1D genomic data. This framework establishes a 12-layer network backbone and parametrizes a long-short term memory network (LSTM) as a controller to search for suitable 1D operations and residual connections, following the ENAS \citep{pham2018enas} optimization protocol. At each step, the controller samples architectures to compute reward based on area under the receiver operating characteristics (AUROC) before outputting the highest-reward architecture. We evaluate AMBER on the ECG, Satellite, and DeepSEA tasks. 

\paragraph{General-Purpose Baselines: Perceiver IO and XGBoost}
As the overarching theme of NAS-Bench-360 is evaluate NAS methods on a wide variety of diverse tasks and when to even use NAS methods over fixed baselines, general-purpose non-NAS methods are obvious points of comparison. 
We evaluate two such baselines on NAS-Bench-360: the recent transformer-based Perceiver IO \citep{jaegle2022perceiver}, and the popular non-deep learning baseline, XGBoost \citep{xgboost}. Perceiver IO is a general-purpose transformer architecture that is designed to handle arbitrary input and output dimensionalities with minimal changes to its encoder and decoder networks---as such, we evaluate Perceiver IO on all 10 NAS-Bench-360 tasks. We note that Perceiver IO tends to perform best in settings with more data than in present in the NAS-Bench-360 tasks. 
Similarly, the popular gradient-boosting method, XGBoost, is applicable to a wide variety of tasks and learning objectives, including single-output and multi-output classification and regression problems, which covers all 10 tasks in NAS-Bench-360. For efficiency and comparison to deep learning methods, we employ the GPU-based implementation of histogram gradient-boosting in XGBoost.

\section{Comparison of NAS with Expert Architectures}
\begin{figure}[!t]
	
	\centering
	\includegraphics[width=0.99\linewidth]{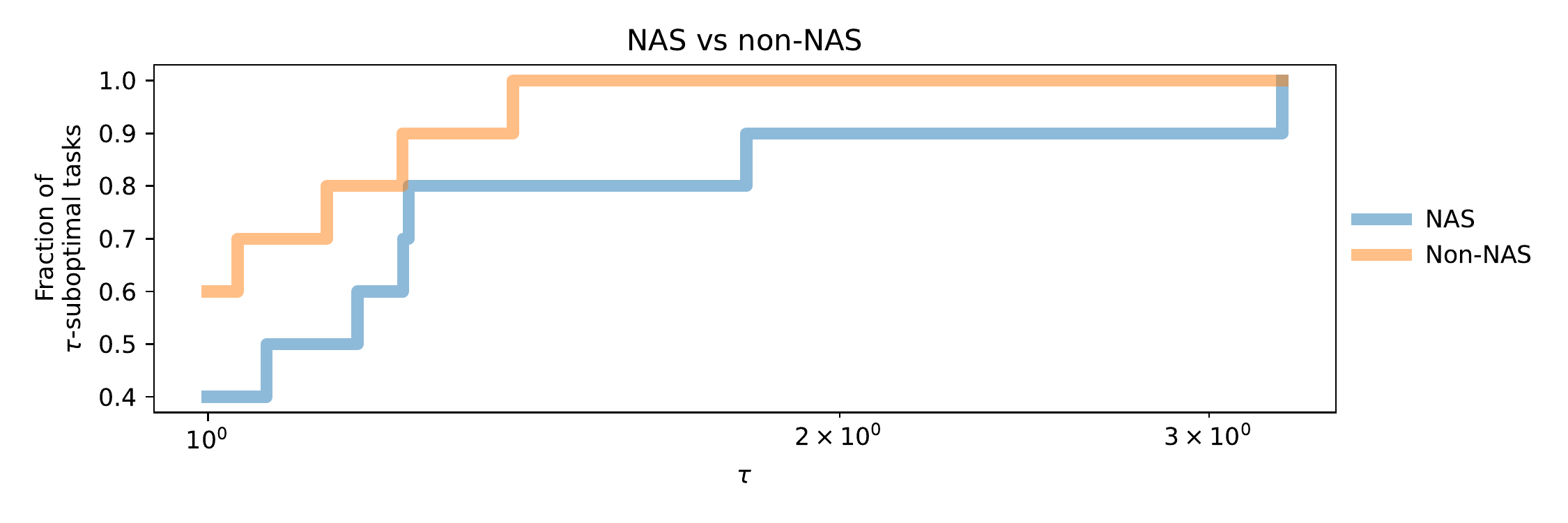}
	\caption{\label{fig:nasvnonnas}
		Performance profiles on all tasks for best-performing NAS vs. Non-NAS.
		The y-value indicates the fraction of tasks on which a plotted method’s error is within a multiplicative factor $\tau$ of the lowest error achieved by all plotted methods..
	}
	
\end{figure}

Hand-crafted networks are selected according to best-effort search. 
In Table~\ref{table-2}, we compare all NAS methods to these hand-crafted networks, denoted as expert architectures. 
Figure \ref{fig:nasvnonnas} illustrates a comparison between best-performing NAS methods vs. best non-NAS methods. Surprisingly, GAEA PC-DARTS beats all the baselines on a portion of the tasks. 

Here is a brief summary of these expert models and their citations:
\begin{enumerate}

\item DenseNet-BC (CIFAR-100): a more sophisticated version of ResNet, achieving state-of-the-art performance on vision classification \citep{huang2017densely}. 

\item S2CNN (Spherical): a spherical CNN contains special operations designed for spherical signals, state-of-the-art on spherically-projected MNIST \citep{cohen2018spherical}. 

\item Fourier Neural Operator (FNO) Network (Darcy Flow): via parametrization in Fourier space, FNO can efficiently learn a family of partial differential equations and their mapping to solutions \citep{li2021fno}. 

\item DEEPCON (PSICOV): a dilated-convolution neural network combined with dropout to optimize for protein distance prediction \citep{adhikari2020deepcon}.

\item deepCR-mask (Cosmic): a modified version of UNet retaining data dimension to keep pixels at the borders to suit astronomy applications, state-of-the-art on this task \citep{zhang2020deepcr}. 

\item Attention-based model (NinaPro): a lightweight feed-forward neural network adopting attention modules in place of convolutions \citep{josephs2020semg}. 

\item VGG-like (FSD50K): a smaller VGG network with output features combining both global max pooling and average pooling for audio \citep{fonseca2020fsd50k}. 

\item ResNet-1D (ECG): ResNet with 1D convolution, using a larger kernel size of 16 and a stride of 2 for all convolutions. The architecture is state-of-the-art on several time-series prediction tasks in medicine \citep{hong2020holmes}. 

\item ROCKET (Satellite): a simple linear classifier with random convolution kernel as a feature extractor, achieving state-of-the-art performance on UCR time-series prediction tasks \citep{dempster2020rocket}. 

\item DeepSEA model (DeepSEA): the original 1D convolution model accompanying the dataset, state-of-the-art on DeepSEA itself \citep{zhou2015predicting}. 

\end{enumerate}

\section{Experiment Details}
\vspace{1mm}

\subsection{Hyperparameter Tuning and Backbone}
We use a wide residual network with 16 layers and a widening factor of 4 (WRN-16-4) for all tasks. 

For tuning hyperparameters, we use ASHA's default elimination schedule and search over 7 to 256 randomly sampled
hyperparameter configurations matching GAEA PC-DART's runtime. The maximum epochs that a single configuration could be trained is equal to that of Wide ResNet's default, 200.

We have selected the following hyperparameter ranges for tuning the Wide ResNet backbone:
\begin{itemize}
	\item $\log_{10}$(learning rate): Unif[-4, -1]
	\item  momentum: Unif\{0.0, 0.3, 0.6, 0.9\}
	\item $\log_{10}$(weight decay): Unif[-5, -2]
	\item dropout: Unif\{0.0, 0.3, 0.6\}
	\item batch size: 128 (all point tasks except FSD50K), 4 (Darcy Flow), 8 (PSICOV, Cosmic), 256(FSD50K, ECG, Deepsea), 4096 (Satellite)
\end{itemize}



\subsection{Reference Runtimes}
Using a Nvidia V100 GPU with 16GB of memory, we have recorded the following runtimes for each experiment in this benchmark in Table \ref{table-4}. 
Overall, GAEA PC-DARTS is more costly than backbone with hyperparameter optimization, which is more costly than DenseNAS. The protein tasks requires heavy computation since the data is not static but generated during training. 

\begin{table}
	\caption{Experiment training runtimes of NAS-Bench-360 (GPU hours)}
	\label{table-4}
	\centering
	\begin{tabular}{lcccc}
		\toprule
		
		Task     &  GAEA  &  DenseNAS &  WRN & AMBER / Auto-DeepLab\\ 
		\midrule
		CIFAR-100 & 9.5 & 2.5& 2& n/a\\
		\midrule 
		Spherical  & 16.5 & 2.5& 2& n/a\\
		\midrule
		Darcy Flow & 6.5 & 0.5& 0.5 &5.5\\ 
		\midrule
		PSICOV & 18 &24& 18.5& 19\\ 
		\midrule 
		Cosmic & 21.5& 2.5& 4& 17.5\\ 
		\midrule
		NinaPro  & 0.5 & 0.2& 0.2& n/a\\ 
		\midrule 
		FSD50K &37&4.5&4& n/a \\
		\midrule 
		ECG &140&6.5&5&27 \\
		\midrule 
		Satellite &28&3&4.5&26 \\
		\midrule 
		DeepSEA &39.5&2&1.5&28 \\
			\bottomrule
	\end{tabular}
\end{table}

\subsection{Model Sizes and FLOPS Statistics}

Full information of model parameter counts and FLOPs can be found in Table~\ref{table-6} and Table~\ref{table-7}.

\subsection{Adjustments for Dense Prediction Tasks}

On the wide ResNet backbone, we add an adaptive averaging pooling operation to upsample the features back to their original dimensions before output.  On the DARTS space, we prevent downsampling and keep spatial dimensions unchanged by disabling reduction cells and replacing them with normal cells. On DenseNAS, we configure the super-network to contain only blocks with the original spatial dimensions. 

\subsection{Adjustments for 1D Prediction Tasks}
The WRN-1D does not have a convolution stem and uses larger kernel sizes of 8, 5, 3 in each convolution block. We substitute 2D operations with 1D operations within the DARTS and DenseNAS search spaces.

\subsection{Random Seeds}
For main experiments, we fix the random seed to be 0, 1, 2 for each of the 3 trials respectively.

For AMBER experiments, we completed three trials as the package did not offer the option of setting random seeds. 

\subsection{Correlation Between Performance and Model Size}

\begin{figure}[!t] 
\includegraphics[width=\linewidth]{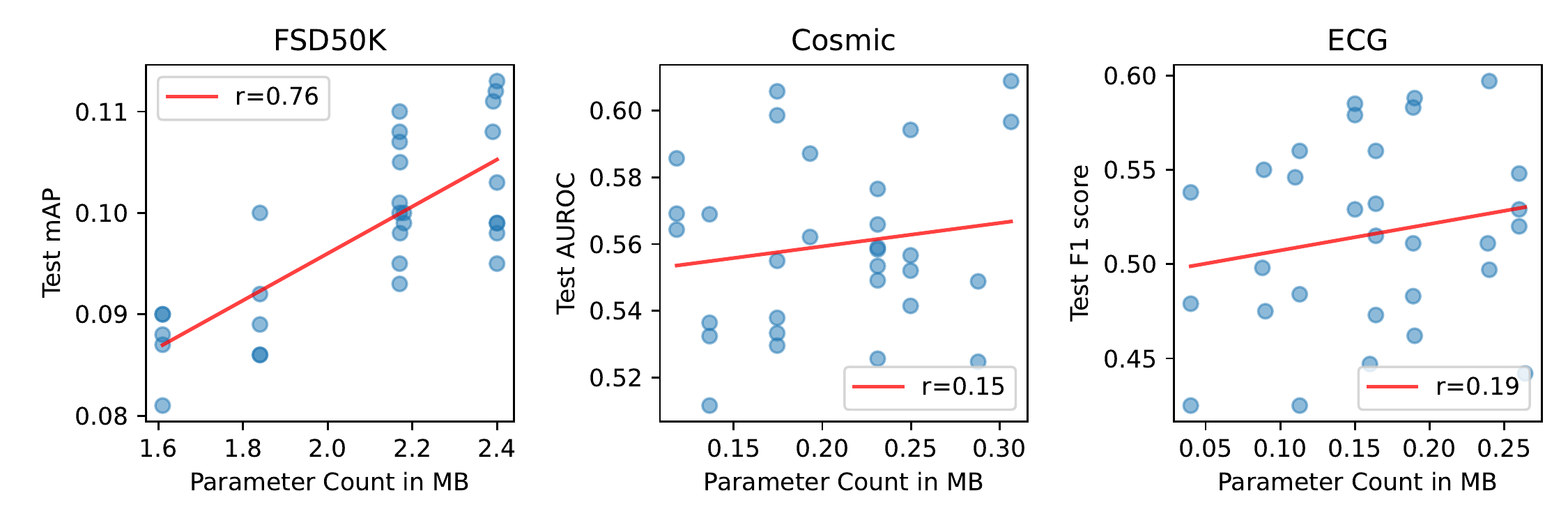}
\caption{Performances by model sizes for three sample tasks.}
\label{fig:random}
\end{figure}

We plot performances of 30 random architectures from the DenseNAS search space across three tasks in Figure \ref{fig:random}. From our random search experiment, larger models searched by NAS are not always better-performing. We study the Pearson correlation coefficient between test performance vs. model size in number of parameters for three tasks: FSD50K, Cosmic, and ECG. On Cosmic and ECG, the correlation is very weak ($r=0.15$ and $r=0.19$ respectively). On FSD50K, a stronger correlation ($r=0.79$) is observed but performance varies significantly even for architectures of the same size.


\subsection{Internal Validation of Dense Task Diversity}
We conduct an internal validation of the diversity of a subset of the NAS-Bench-360 tasks, namely the dense prediction tasks Darcy Flow, PSICOV, and Cosmic. We compare the expert architectures for each task to the expert architecture for Darcy Flow: the FNO-based architecture. 
The results of this study are shown in Table~\ref{table:dense-validation}. We observe a drop in performance using the expert FNO architecture on our three dense tasks. FNO was designed for Darcy Flow and achieves the best performance on this task across all methods. In contrast, we find that FNO performance is considerably worse on the other two dense tasks, PSICOV and Cosmic.

\begin{table}
	\centering
	\begin{threeparttable}
		\caption{
			Performance of the expert-designed FNO architecture (designed for Darcy Flow) across all dense tasks.\looseness-1
		}
		\label{table:dense-validation}
		
		\begin{tabular}{lccc}
			\toprule
			                                & Darcy Flow & PSICOV & Cosmic \\
			\midrule
			Expert (Darcy Flow) & {\bf \res{0.008}{0.001}} & \res{4.43}{0.055}        & \res{0.301}{0.01} \\
			Expert                      & {\bf \res{0.008}{0.001}} & {\bf \res{3.35}{0.140}} & {\bf \res{0.127}{0.01}} \\
			\bottomrule
		\end{tabular}
		
		\footnotesize

	\end{threeparttable}
\end{table}

\section{Dataset Details}
\vspace{1mm}


\subsection{Data License}
\begin{itemize}
	\item CIFAR-100: CC BY 4.0 (on \url{https://www.tensorflow.org/datasets/catalog/cifar100})
	\item Spherical CIFAR-100: CC BY-SA
	\item NinaPro: CC BY-ND
	\item FSD50k: CC BY 4.0
	\item Darcy Flow: MIT
	\item DeepCov, PSICOV: GPL
	\item Cosmic: CC BY 4.0
	\item ECG: ODC-BY 1.0 
	\item Satellite: GPL 3.0
	\item Deepsea: CC BY 4.0
\end{itemize}

\subsection{Data Preprocessing Details} 
\paragraph{CIFAR-100:} while the 10,000 testing images are kept aside only for evaluating architectures, the 50,000 training images are randomly partitioned into 40,000 for architecture search and 10,000 for validation. On all of the 50,000 training images, we apply standard CIFAR augmentations including random crops and horizontal flipping, and finally normalize them using a pre-calculated mean and standard deviation of this set. On the 10,000 testing images, we only apply normalization with the same constants. 

\paragraph{Spherical:}  with the same split ratios CIFAR-100, the generated spherical image data is directly used for training and evaluation without data augmentation and pre-processing.

\paragraph{NinaPro:} Containing less than 4,000 samples, the data is comprised of single-channel signals with an irregular shape of 16*52 pixels. This task also differs from CIFAR for its class imbalance, as over $65\%$ of all gestures are the neutral position. We split the data using the same ratio as CIFAR, resulting in 2638 samples for training, and 659 samples for validation and testing each. No additional pre-processing is performed. 

\paragraph{FSD50K:} The raw sound files are first resampled at a frequency of 22,050 Hz and transformed into 96-band, log-mel spectrograms, which is a representation of the sound's power spectrum. Small overlapping audio chunks of 1 second are obtained from these larger clips, resulting in an input size of 101*96 (101 frames of 96-band spectrograms). As data augmentation, background noise of the same frequency is also mixed into 75\% of the training data. We split 4,170 clips into the validation set and 10,231 clips into the test set following the original paper. During training, we train on one randomly-sampled chunk, instead of all chunks, from each clip. 

\paragraph{Darcy Flow:} we use scripts provided by \citep{li2021fno} to generate the PDEs and their solutions, for a total of 900 data points for training, 100 for validation, and 100 for testing. All input data is normalized with constants calculated on the training set before fed into the neural network and de-normalized following an encode-decode scheme. The solutions, or labels, for the training set are also encoded and decoded this way. The test labels are not processed. 

\paragraph{PSICOV:} we adopt the chosen subset of DeepCov proteins in \citep{adhikari2020fully}, consisting of 3,456 proteins each with 128*128 feature maps across 57 channels. 100 proteins from this set are used for validation and the rest for training. Test data for final evaluation is gathered from another set of 150 proteins, PSICOV. Since these produce feature maps that are larger (512*512), we run the prediction network over all of its non-overlapping 128*128 patches. 

\paragraph{Cosmic:} we use data from a specific filter, F435W, of the space telescope, representing the 3605–4882 \text{\normalfont\AA} spectral range. Image stamps of 256*256 pixels are taken from large images. The dataset contains 4,347 stamps for training, and 420 for test, and 483 for validation to match the test set size. 

\paragraph{ECG:} from the sliding window approach, 12,186 single lead recordings are converted into more than 330,000 recording segments comprised of 261,740 for training, 33,281 for validation, and 33,494 for test. Each segment is of the shape 1*1,000, representing one channel of 1,000-long temporal sequence. 

\paragraph{Satellite:} each satellite time-series is single-channel of length 46 (1*46). After applying standard normalization, we divide the one million entries to 800,000 for training, 100,000 for validation, and 100,000 for test. Zero-padding to 48-length sequences is required for DenseNAS' downsampling network.

\paragraph{DeepSEA:} the genome sequences are 1,000-base pair (bp) long and represented as a 1000$\times$4 binary matrix, as each bp is represented as an one-hot encoding corresponding to either A,C,T,G at that location. Total training set size is 71,753. Validation and test sizes that are not subsampled are 2,490 and 149,400 respectively.

\begin{table}
	\centering
	\begin{threeparttable}
		\caption{
			Paramater counts of searched and baseline models for all tasks of NAS-Bench-360. 
			Searched model sizes are reported as \res{mean}{standard deviation} of three random seeds. Results are reported in millions (M). Architectures with the best performance are bolded.
		}
		\label{table-6}

		\footnotesize
		\begin{tabular}{llccccc}
			\toprule
			Search space & Search algo. & CIFAR-100 & Spherical  & Darcy Flow & PSICOV & Cosmic \\

			\midrule
			DenseNAS & random & \res{1.74}{0.12}& \res{2.23}{0.47}&\res{\,\,\,1.00}{0.18}& \res{1.21}{0.16}& \res{0.25}{0.06}  \\
			
			DenseNAS & original & \res{2.03}{0.53} &\res{1.84}{0.15} & \res{\,\,\,0.38}{0.13}& \res{0.93}{0.36}& \res{0.15}{0.16} \\
			
			DARTS & GAEA &\res{4.92}{0.28}  & \textbf{\res{1.67}{0.14}}  & \res{\,\,\,0.63}{0.08}  & \textbf{\res{0.53}{0.05}} & \res{0.43}{0.15}  \\  	
				
			Auto-DL & DARTS &n/a& n/a&\res{22.98}{3.49}&\res{6.50}{1.84}& \res{7.61}{2.14}\\

			\midrule 
			WRN & default &2.77& 2.77& 2.75& 2.76& 2.75 \\
			Expert & default & \textbf{3.08} & 0.16 &\textbf{1.19} & 0.60 & \textbf{0.10} \\
			
			\toprule
			\toprule
			Search space & Search algo. & NinaPro & FSD50K  & ECG & Satellite & DeepSEA \\

			\midrule
			
			DenseNAS & random & \res{6.80}{0.46}&\textbf{\res{2.40}{0.00}}&\res{0.18}{0.05}&\res{0.79}{0.16}&\res{0.25}{0.04} \\
			
			DenseNAS & original & \res{6.69}{0.53}&\res{1.45}{0.00}&\res{0.11}{0.05}&\res{1.08}{0.63}&\res{0.19}{0.00} \\
			
			DARTS & GAEA & \res{3.35}{0.48} & \res{0.81}{0.11} & \res{3.31}{0.07}& \textbf{\res{3.35}{0.35}} &  \res{2.91}{0.47}\\ 
			
			AMBER & ENAS&n/a& n/a& \res{6.61}{0.33}&\res{6.22}{1.36}& \res{8.44}{1.47}\\

			\midrule
			WRN & default & \textbf{2.75}&2.80&0.50&0.51& 0.51 \\
			Expert & default & 1.36 & 0.35& \textbf{16.5} & 0.48 &\textbf{60.9} \\
			
			\bottomrule
		\end{tabular}
	\end{threeparttable}
\end{table}

\begin{table}
	\centering
	\begin{threeparttable}
		\caption{
			FLOPS of searched and baseline models for all tasks of NAS-Bench-360. 
			Searched model FLOPS are reported as \res{mean}{standard deviation} of three random seeds. Results are reported in GFLOPS. Architectures with the best performance are bolded.
		}
		\label{table-7}
		
		\footnotesize
		\begin{tabular}{llccccc}
			\toprule
			Search space & Search algo. & CIFAR-100 & Spherical  & Darcy Flow & PSICOV & Cosmic \\
			
			\midrule
			DenseNAS & random & \res{0.46}{0.07}& \res{0.91}{0.07}&\res{14.42}{2.58}& \res{39.80}{5.09}& \res{\,\,\,8.42}{2.11}  \\
			
			DenseNAS & original & \res{0.44}{0.53} &\res{1.84}{0.15} & \res{\,\,\,5.43}{1.82}& \res{\,\,\,30.51}{11.90}& \res{\,\,\,5.00}{5.30} \\
			
			DARTS & GAEA &\res{1.42}{0.09}  & \textbf{\res{1.91}{0.65}}  & \res{\,\,\,9.33}{1.13}  & \textbf{\res{17.74}{1.68}} & \res{14.27}{4.90}  \\  	
			
			Auto-DL & DARTS &n/a& n/a&\res{\,\,\,2.54}{1.20}&\res{\,\,\,3.43}{1.27}& \res{\,\,\,2.44}{0.26}\\

			\midrule 
			WRN & default &0.78& 2.78& 39.72& 90.58& 90.06 \\
			Expert & default & \textbf{1.18} & n/a &\textbf{n/a} & 0.01 & \textbf{1.96} \\
			
			\toprule
			\toprule
			Search space & Search algo. & NinaPro & FSD50K  & ECG & Satellite & DeepSEA \\

			\midrule
			
			DenseNAS & random & \res{1.02}{0.06}&\textbf{\res{	0.40}{0.00}}&\res{0.11}{0.02}&\res{0.02}{0.01}&\res{0.15}{0.02} \\
			
			DenseNAS & original & \res{0.97}{0.14}&\res{\,0.80}{0.00}&\res{0.16}{0.03}&\res{0.02}{0.01}&\res{0.10}{0.00} \\
			
			DARTS & GAEA & \res{0.89}{0.12} & \res{\,2.57}{0.47} & \res{2.28}{0.05}& \textbf{\res{0.11}{0.07}} &  \res{2.01}{0.33}\\ 
			
			AMBER & ENAS&n/a& n/a& \res{0.03}{0.01}&\res{0.03}{0.01}& \res{0.04}{0.01}\\

			\midrule
			WRN & default & \textbf{0.64}&7.56&1.02&0.04& 1.02 \\
			Expert & default & 0.02 & 0.66& \textbf{0.70} & 0.01 &\textbf{0.12} \\
			
			\bottomrule
		\end{tabular}
		\begin{tablenotes}
		\item *some expert models contain non-standard modules without FLOPS count.
	\end{tablenotes}
	\end{threeparttable}
\end{table}

\section{Ethics and Responsible Use}\label{app:ethics}

Within our array of tasks, only NinaPro, ECG, and DeepSEA contain human-derived data. Our chosen subset of NinaPro contains only muscle movement data without any exposure of personal information. The original experiments to acquire NinaPro data are approved by the ethics commission of the canton of Valais, Switzerland \citep{atzori2012building}. The ECG data derives from an open challenge and is provided by the medical device company AliveCor, under the GPL license allowing it for public use. The DeepSEA data derived from ENCODE is part of an international collaborative effort, which is overseen and funded by the National Human Genome Research Institute (NHGRI). 
For other datasets, we have listed the data licenses in the appendix.
While we do not view the specific datasets in NAS-Bench-360 as potential candidates for misuse, the broader goal of applying NAS to new domains comes with inherent risks that may require mitigation on an application-by-application basis.

\end{document}